%% file: main.tex
\documentclass{article}

\usepackage[utf8]{inputenc}
\usepackage[margin=1.5in]{geometry}

\usepackage{natbib}
\usepackage{graphicx}
\usepackage{tabularx}
\usepackage{placeins}
\usepackage[toc,page]{appendix}

\usepackage[utf8]{inputenc} 
\usepackage[T1]{fontenc}    
\usepackage{hyperref}       
\hypersetup{ %
    pdfborder=0 0 0,
    pdfpagemode=UseNone,
    colorlinks=true,
    linkcolor=blue,
    citecolor=blue,
    filecolor=blue,
    urlcolor=blue,
    pdfview=FitH}
    
\usepackage{url}            
\usepackage{booktabs}       
\usepackage{amsfonts}       
\usepackage{nicefrac}       
\usepackage{microtype}      
\usepackage{color}
\usepackage{amsmath}

\usepackage{ifthen}         
\usepackage{boldline}       

\usepackage[font=small,labelfont=bf]{caption}
\usepackage[font=small,labelfont=bf]{subcaption}
\usepackage{verbatim}
\usepackage{xpatch}
\usepackage{setspace}
\usepackage{adjustbox}

\usepackage{algorithm}
\usepackage[noend]{algpseudocode}

\makeatletter
\xpatchcmd{\algorithmic}{\itemsep\z@}{\itemsep=1ex plus2pt}{}{}  
\makeatother

\usepackage{multirow}
\usepackage{arydshln}
\usepackage{rotating}
\usepackage{array}
\usepackage{mfirstuc}
\usepackage[dvipsnames]{xcolor}

\input{macros.tex}
\title{Spatial Broadcast Decoder: A Simple Architecture for Learning Disentangled Representations in VAEs}

\author{
  Nicholas Watters,\; Loic Matthey,\; Christopher P. Burgess, Alexander Lerchner\\
  DeepMind\\
  London, United Kingdom\\
  \texttt{\{nwatters, lmatthey, cpburgess, lerchner\}@google.com}\\
}

\date{}

\begin{document}

\maketitle

\begin{abstract}
We present a simple neural rendering architecture that helps variational autoencoders (VAEs) learn disentangled representations.
Instead of the deconvolutional network typically used in the decoder of VAEs, we tile (broadcast) the latent vector across space, concatenate fixed X- and Y-``coordinate'' channels, and apply a fully convolutional network with $1\times 1$ stride.
This provides an architectural prior for dissociating positional from non-positional features in the latent distribution of VAEs, yet without providing any explicit supervision to this effect.
We show that this architecture, which we term the \emph{Spatial Broadcast decoder}, improves disentangling, reconstruction accuracy, and generalization to held-out regions in data space.
It provides a particularly dramatic benefit when applied to datasets with small objects.
We also emphasize a method for visualizing learned latent spaces that helped us diagnose our models and may prove useful for others aiming to assess data representations.
Finally, we show the Spatial Broadcast Decoder is complementary to state-of-the-art (SOTA) disentangling techniques and when incorporated improves their performance.
\end{abstract}

\input{introduction.tex}

\input{model.tex}

\input{related_work.tex}

\input{results.tex}

\input{conclusion.tex}

\vspace*{0.4in}

\subsubsection*{Acknowledgments}

We thank Irina Higgins, Danilo Rezende, Matt Botvinick, and Yotam Doron for helpful discussions and insights.

\FloatBarrier
\clearpage

\bibliographystyle{plainnat}
\bibliography{references}

\newpage

\input{supplementary.tex}

\end{document}

%% file: macros.tex
\usepackage{amsmath,amsthm,amssymb}
\usepackage{xspace}

\newcommand\cut[1]{}

\newcommand{\betavae}{\ensuremath{\beta}-VAE\xspace}

\newcolumntype{C}[1]{>{\centering\arraybackslash}m{#1}}
\newcolumntype{R}[1]{>{\raggedleft\arraybackslash}m{#1}}

\newcommand{\myvec}[1]{\mathbf{#1}}

\newcommand{\vx}{\myvec{x}}

\newcommand{\vy}{\myvec{y}}

\newcommand{\vz}{\myvec{z}}

\newcommand{\be}{\begin{equation}}
\newcommand{\ee}{\end{equation}}
\newcommand{\bea}{\begin{eqnarray}}
\newcommand{\eea}{\end{eqnarray}}
\newcommand{\beaa}{\begin{eqnarray*}}
\newcommand{\eeaa}{\end{eqnarray*}}

\DeclareMathAlphabet{\mathpzc}{OT1}{pzc}{m}{n}

%% file: introduction.tex
\section{Introduction}\label{S:intro}

Knowledge transfer and generalization are hallmarks of human intelligence.
From grammatical generalization when learning a new language to visual generalization when interpreting a Picasso, humans have an extreme ability to recognize and apply learned patterns in new contexts.
Current machine learning algorithms pale in contrast, suffering  from overfitting, adversarial attacks, and domain specialization \citep{Lake_etal_2016, marcus_2018}.
We believe that one fruitful approach to improve generalization in machine learning is to learn compositional representations in an unsupervised manner.
A compositional representation consists of components that can be recombined, and such recombination underlies generalization.
For example, consider a pink elephant.
With a representation that composes color and object independently, imagining a pink elephant is trivial.
However, a pink elephant may not be within the scope of a representation that mixes color and object.
Compositionality comes in a variety of flavors, including feature compositionality (e.g. pink elephant), multi-object compositionality (e.g. elephant next to a penguin), and relational compositionality (e.g. the smallest elephant).
In this paper we will focus on feature compositionality.

Representations with feature compositionality are sometimes referred to as ``disentangled'' representations \citep{Bengio_etal_2013}.
However, there is as yet no consensus on the definition of a disentangled representation \citep{higgins2018, Locatello_etal_2018, higgins2017}.
In this paper, when evaluating disentangled representations we both employ standard metrics \citep{Kim_Mnih_2017,chen_2018, Locatello_etal_2018} and (in view of their limitations) emphasize visualization analysis.

\begin{figure}[t!]
  \centering
  \begin{minipage}[t]{.43\linewidth}
    \vspace{0pt} 
    \includegraphics[width=1\linewidth]{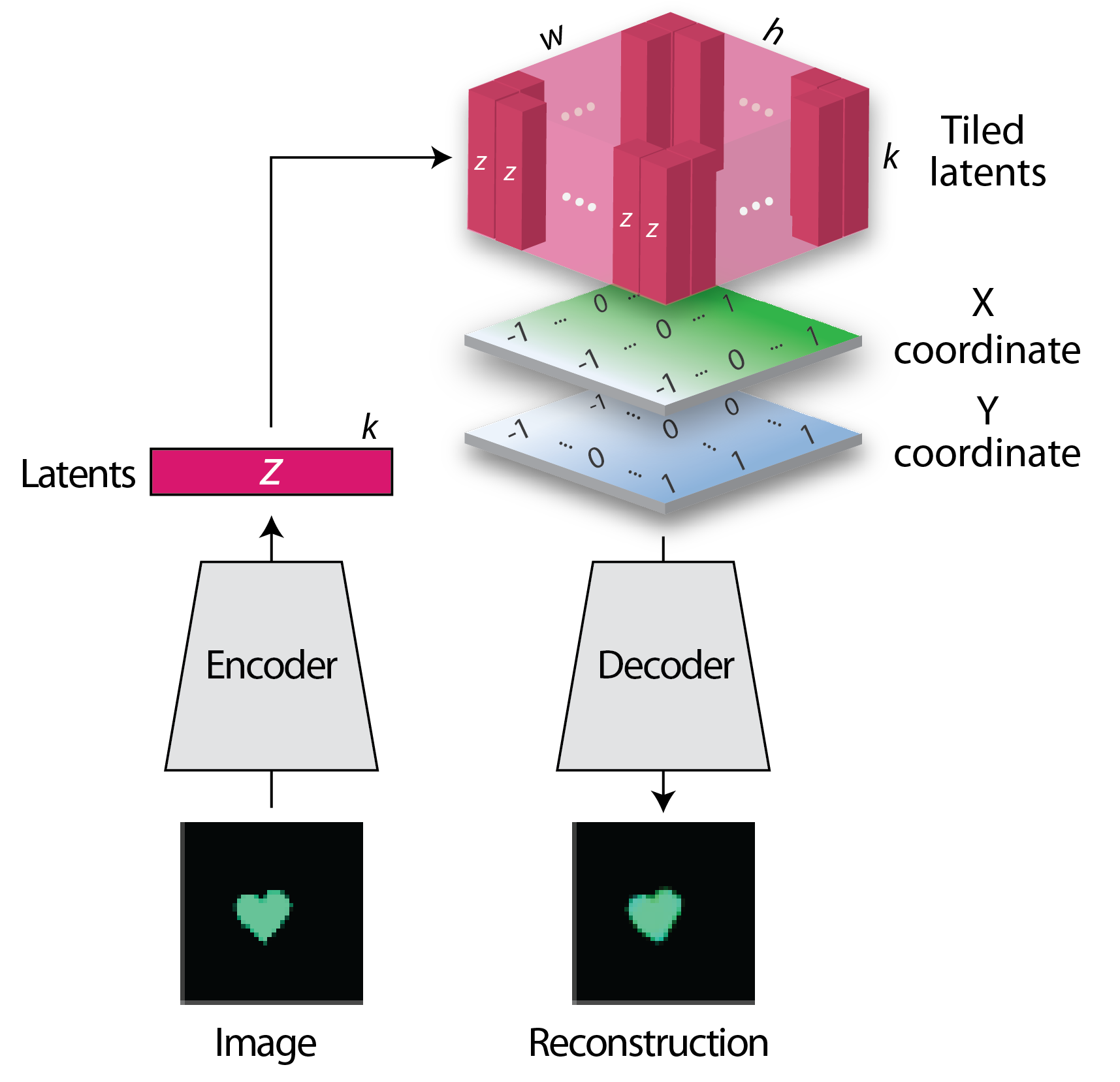}
    \vspace{0pt}
  \end{minipage}
  \hspace{0pt}
  \begin{minipage}[t]{.47\linewidth}
    \vspace{0pt}
    \begin{algorithm}[H]
      \begin{algorithmic}[1]
        \Require{latents $\vz \in \mathcal{R}^k$, width $w$, height $h$}
        \Ensure{tiled latents $\vz_{sb} \in \mathcal{R}^{h \times w \times (k+2)}$}
        \caption{Spatial Broadcast}\label{alg:spatial_broadcast}
            \State $\vz_b = \textproc{tile}(\vz, (h, w, 1))$
            \State $\vx = \textproc{linspace}(-1, 1, w)$
            \State $\vy = \textproc{linspace}(-1, 1, w)$
            \State $\vx_b, \vy_b = \textproc{meshgrid}(\vx, \vy)$
            \State $\vz_{sb} = \textproc{concat}([\vz_b, \vx_b, \vy_b], \text{axis}=-1)$
            \State \Return $\vz_{sb}$
      \end{algorithmic}
     \end{algorithm}
     \vspace{0pt}
  \end{minipage}%
  \vspace{-15pt}
  \caption{\emph{(left)} Schematic of the Spatial Broadcast VAE. In the decoder, we broadcast (tile) a latent sample of size $k$ to the image width $w$ and height $h$, and concatenate two ``coordinate'' channels. This is then fed to an unstrided convolutional decoder. \emph{(right)} Pseudo-code of the spatial broadcast operation, assuming access to a \texttt{numpy} / Tensorflow-like API.}\label{fig:schematic}
  \vspace{-10pt}
\end{figure}

Learning disentangled representations from images has recently garnered much attention.
However, even in the best understood conditions, finding hyperparameters to robustly obtain such representations still proves quite challenging \citep{Locatello_etal_2018}.
Here we present a simple modification of the variational autoencoder (VAE) \citep{Kingma_Welling_2014, Rezende_etal_2014} decoder architecture that can dramatically improve both the robustness to hyperparameters and the quality of the learned representations in a variety of VAE-based models.
We call this the Spatial Broadcast decoder, which we propose as an alternative to the standard MLP/deconvolutional network (which we refer to as the DeConv decoder).
See Figure \ref{fig:schematic} for a schematic.

In this paper we show benefits of using the Spatial Broadcast decoder for image representation-learning, including:
\begin{itemize}
    \item Improvements to both disentangling and reconstruction accuracy on datasets of simple objects.
    \item Complementarity to (and improvement of) state-of-the-art disentangling techniques.
    \item Particularly significant benefits when the objects in the dataset are small, a regime that is notoriously difficult for standard VAE architectures.
    \item Improved representational generalization to out-of-distribution test datasets involving both interpolation and extrapolation in latent space.
\end{itemize}

We also introduce a simple method for visualizing latent space geometry that we found helpful for gaining better insights when qualitatively assessing models and may prove useful to others interested in analyzing autoencoder representations

%% file: model.tex
\section{Spatial Broadcast Decoder}\label{S:model}

When modeling the distribution of images in a dataset with a variational autoencoder \citep{Kingma_Welling_2014, Rezende_etal_2014}, standard architectures use an encoder consisting of a downsampling convolutional network followed by an MLP and a decoder consisting of an MLP followed by an upsampling deconvolutional network.
The convolutional and deconvolutional networks share features across space, improving training efficiency for VAEs much like they do for all models that use image data.

However, while convolution surely lends some useful spatial inductive bias to the representations, a standard VAE learns highly entangled representations in an effort to represent the data as closely as possibly to its Gaussian prior (e.g. see Figure \ref{fig:latent_space_geometry}).

A number of new variations of the VAE objective have been developed to alleviate this problem, though all of them introduce additional hyperparameters \citep{higgins2017, burgess2018, Kim_Mnih_2017, chen_2018}.
Furthermore, a recent study found them to be extremely sensitive to these hyperparameters \citep{Locatello_etal_2018}.

Meanwhile, upsampling deconvolutional networks (like the one in the standard VAE's decoder) have been found to pose optimization challenges, such as producing checkerboard artifacts \citep{odena2016} and spatial discontinuities \citep{liu2018}, effects that seem likely to raise problems for representation-learning in the VAE's latent space.

Intuitively, asking a deconvolutional network to render an object at a particular position is a tall order --- the network's filters have no explicit spatial information, in other words they don't ``know where they are.''
Hence the network must learn to propagate spatial asymmetries down from its highest layers and in from the spatial boundaries of the layers.
This requires learning a complicated function, so optimization is difficult.
To remedy this, in the Spatial Broadcast decoder we remove all upsampling deconvolutions from the network, instead tiling (broadcasting) the latent vector across space, appending fixed coordinate channels, then applying an unstrided fully convolutional network.
This operation is depicted and described in Figure \ref{fig:schematic}.
With this architecture, rendering an object at a position becomes a very simple function (essentially just a thresholding operation in addition to the local convolutional features), though the network still has capacity to represent some more complex datasets (e.g. see Figure \ref{fig:chair_objects_room}).
Such simplicity of computation yields ease of optimization, and indeed we find that the Spatial Broadcast decoder greatly improves performance in a variety of VAE-based models.

Note that the Spatial Broadcast decoder does not provide any additional supervision to the generative model.
The model must still learn to encode spatial information in its latent space in order to reconstruct accurately.
The Spatial Broadcast decoder only allows the model to use the encoded spatial information in its latent space very efficiently.

In addition to better disentangling, the Spatial Broadcast VAE can on some datasets yield better reconstructions (see Figure \ref{fig:rate_distortion}), all with shallower networks and fewer parameters than a standard deconvolutional architecture.
However it is worth noting that if the data does not benefit from having access to an absolute coordinate system, the Spatial Broadcast decoder could hurt.
A standard DeConv decoder may in some cases more easily place patterns relative to each other or capture more extended spatial correlations.
As we show in Section~\ref{S:no_position}, we did not find this to be the case in the datasets we explored, but it is still a possible limitation of our model.

While the Spatial Broadcast decoder can be applied to any generative model that renders images from a vector representation, here we only explore its application to VAE models.

%% file: related_work.tex
\section{Related Work}\label{S:related}

Some of the ideas behind the Spatial Broadcast decoder have been explored extensively in many other bodies of work.

The idea of appending coordinate channels to convolutional layers has recently been highlighted (and named CoordConv) in the context of improving positional generalization \citep{liu2018}. However, the CoordConv technique had been used beforehand \citep{zhao_2015,Liang2015, watters_2017,Wojna2017, Perez2017, nash17, Ulyanov2017} and its origin is unclear.
The CoordConv VAE \citep{liu2018} incorporates CoordConv layers into an upsampling deconvolutional network in a VAE (which, as we show in Appendix \ref{S:coordconv_vae}, does not yield representation-learning benefits comparable to those of the Spatial Broadcast decoder).
To our knowledge no prior work has combined CoordConv with spatially tiling a generative model's representation as we do here.

Another line of work that has used coordinate channels extensively is language modeling.
In this context, many models use fixed or learned position embeddings, combined in different ways to the input sequences to help compute translations depending on the sequence context in a differentiable manner \citep{Gu2016, Vaswani2017, Gehring2017, Devlin2018}.

We can draw parallels between the Spatial Broadcast decoder and generative models that learn ``where'' to write to an image \citep{Jaderberg2015, Gregor2015, Reed2016}, but in comparison these models render local image patches whereas we spatially tile a learned latent embedding and render an entire image.
Work by \citet{Dorta2017} explored using a Laplacian Pyramid arrangement for VAEs, where some parts of the latent distributions would have global effects, whereas others would modify fine scale details about the reconstruction.
Similar constraints on the extent of the spatial neighbourhood affected by the latent distribution has also been explored in \cite{Parmar2018}.
Other types of generative models already use convolutional latent distributions \citep{Finn2015, Levine2015, Eslami2018}, unlike the flat vectors we consider here.
In this case the tiling operation is not necessary, but adding coordinate channels might be of help.

%% file: results.tex
\section{Results}\label{S:results}

We present here a select subset of our assessment of the Spatial Broadcast decoder.
Interested readers can refer to Appendices~\ref{S:experiment_details}-\ref{S:circles_geometry} for a more thorough set of observations and comparisons.

\input{sota.tex}

\input{small_objects.tex}

\input{dependent_factors.tex}

%% file: sota.tex
\subsection{Performance on colored sprites}\label{S:csprites}

\begin{figure}[t!]
  \centering
  \vspace*{-15pt}
  \begin{subfigure}[b]{0.3\textwidth}  
  \includegraphics[width=1.0\linewidth]{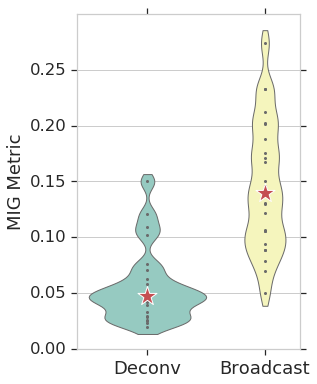}
  \end{subfigure}
  \begin{subfigure}[b]{0.345\textwidth}  
  \includegraphics[width=1.0\linewidth]{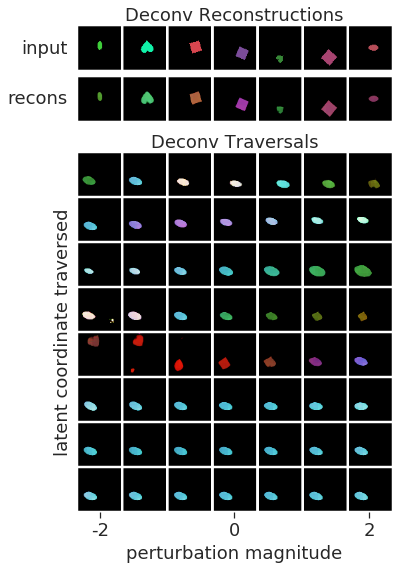}
  \end{subfigure}
  \begin{subfigure}[b]{0.339\textwidth}  
  \includegraphics[width=1.0\linewidth]{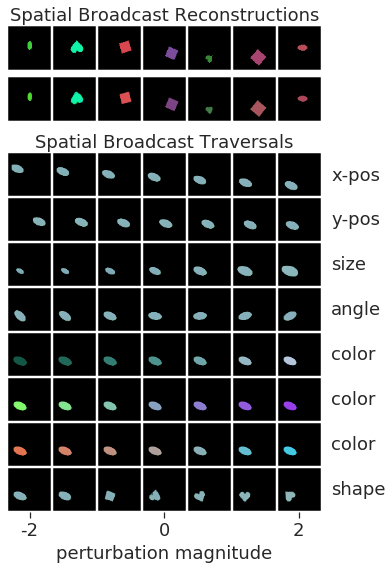}
  \end{subfigure}
  \caption{\textbf{Comparing Deconv to Spatial Broadcast decoder in a VAE.} \emph{(left)} MIG results, showing a Spatial Broadcast VAE achieves higher (better) scores than a DeConv VAE. Stars are median MIG values and the seeds used for the traversals on the right. \emph{(middle)} DeConv VAE reconstructions and latent space traversals. Traversals are generated around a seed point in latent space by reconstructing a sweep from -2 to +2 for each coordinate while keeping all other coordinates constant. The traversal shows an entangled representation in this model. \emph{(right)} Spatial Broadcast VAE reconstructions and traversal. The traversal is well-disentangled and aligned with generative factors, as indicated by the labels on the right (which were attributed by visual inspection). While all models were trained with 10 latent coordinates, only the $8$ lowest-variance ones are shown in the traversals (the remainder are non-coding coordinates).}
  \label{fig:colored_sprites:vae}
  \vspace{-15pt}
\end{figure}

The Spatial Broadcast decoder was designed with object-feature representations in mind, hence to initially showcase its performance we use a dataset of simple objects: colored 2-dimensional sprites.
This dataset is described in the literature \citep{burgess2018} as a colored version of the dSprites dataset \citep{dsprites17}.
One advantage of the colored sprites dataset is it has known factors of variation, of which there are 8:  X-position, Y-position, Size, Shape, Angle, and three-dimensional Color.
Thus we can evaluate disentangling performance with metrics that rely on ground truth factors of variation \citep{chen_2018, Kim_Mnih_2017}.

To quantitatively evaluate disentangling, we focus primarily on the Mutual Information Gap (MIG) metric \citep{chen_2018}.
The MIG metric is defined by first computing the mutual information matrix between the latent distribution means and ground truth factors, then computing the difference between the highest and second-highest elements for each ground truth factor (i.e. the gap), then finally averaging these values over all factors.

Despite significant shortcomings (see Section~\ref{S:latent_geometry}), we did find the MIG metric overall can a helpful tool for evaluating representational quality, though should by no means be trusted blindly. We did also use the FactorVAE metric \citep{Kim_Mnih_2017}, yet found it to be less consistent then MIG (we report these results in Appendix~\ref{S:hyperparams}).

In Figure \ref{fig:colored_sprites:vae} we compare a standard DeConv VAE (a VAE with an MLP + deconvolutional network decoder) to a Spatial Broadcast VAE (a VAE with the Spatial Broadcast decoder). We see that the Spatial Broadcast VAE outperforms the DeConv VAE  both in terms of the MIG metric and traversal visualizations.
The Spatial Broadcast VAE traversals clearly show all 8 factors represented in separate latent factors, seemingly on par with more complicated state-of-the-art methods on this dataset \citep{burgess2018}.
The hyperparameters for the models in Figure \ref{fig:colored_sprites:vae} were chosen over a large sweep to minimize the model's error, not chosen for any disentangling properties explicitly.
See Appendix \ref{S:hyperparams} for more details about hyperparameter choices and sensitivity.

The Spatial Broadcast decoder is complementary to existing disentangling VAE techniques, hence improves not only a vanilla VAE but SOTA models as well.
To demonstrate this, we consider two recently developed models, FactorVAE \citep{Kim_Mnih_2017} and \betavae \citep{higgins2017}.

\begin{figure}[t!]
  \centering
  \begin{subfigure}[b]{0.28\textwidth}
  \includegraphics[width=1.0\linewidth]{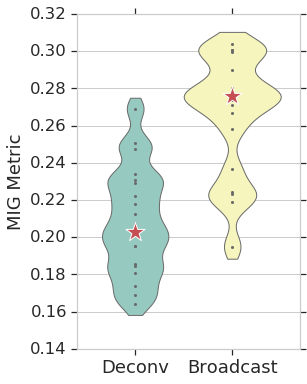}
  \end{subfigure}
  \begin{subfigure}[b]{0.325\textwidth}
  \includegraphics[width=1.0\linewidth]{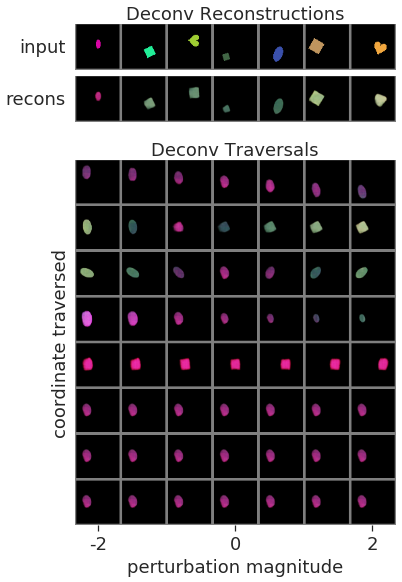}
  \end{subfigure}
  \begin{subfigure}[b]{0.375\textwidth}
  \includegraphics[width=1.0\linewidth]{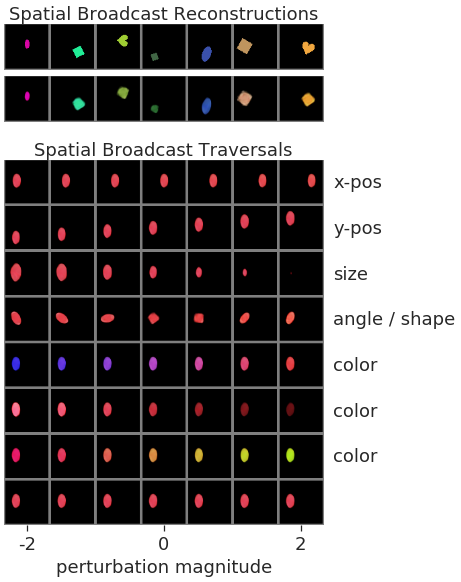}
  \end{subfigure}
  \caption{\textbf{Comparing Deconv to Spatial Broadcast decoder in a FactorVAE.} \emph{(left)} MIG results, showing a Spatial Broadcast FactorVAE acheives higher (better) scores than a  DeConv FactorVAE. Stars are median MIG values and the seeds used for the traversals on the right. \emph{(middle)} DeConv FactorVAE reconstructions and entangled latent space traversals. \emph{(right)} Spatial Broadcast FactorVAE reconstructions and traversal. The traversal is well-disentangled. As in Figure \ref{fig:colored_sprites:vae}, only the most relevant 8 of each model's 10 latent coordinates are shown in the traversals.}
  \label{fig:colored_sprites:factorvae}
\end{figure}

Figure \ref{fig:colored_sprites:factorvae} shows the Spatial Broadcast decoder improving the disentangling of FactorVAE in terms of both the MIG metric and traversal visualizations. See Appendix \ref{S:circles_geometry} for further results to this effect.

\begin{figure}[t!]
  \vspace*{-10pt}
  \centering
  \begin{subfigure}{0.49\textwidth}
  \caption{Reconstruction vs KL}
  \includegraphics[width=1.0\linewidth]{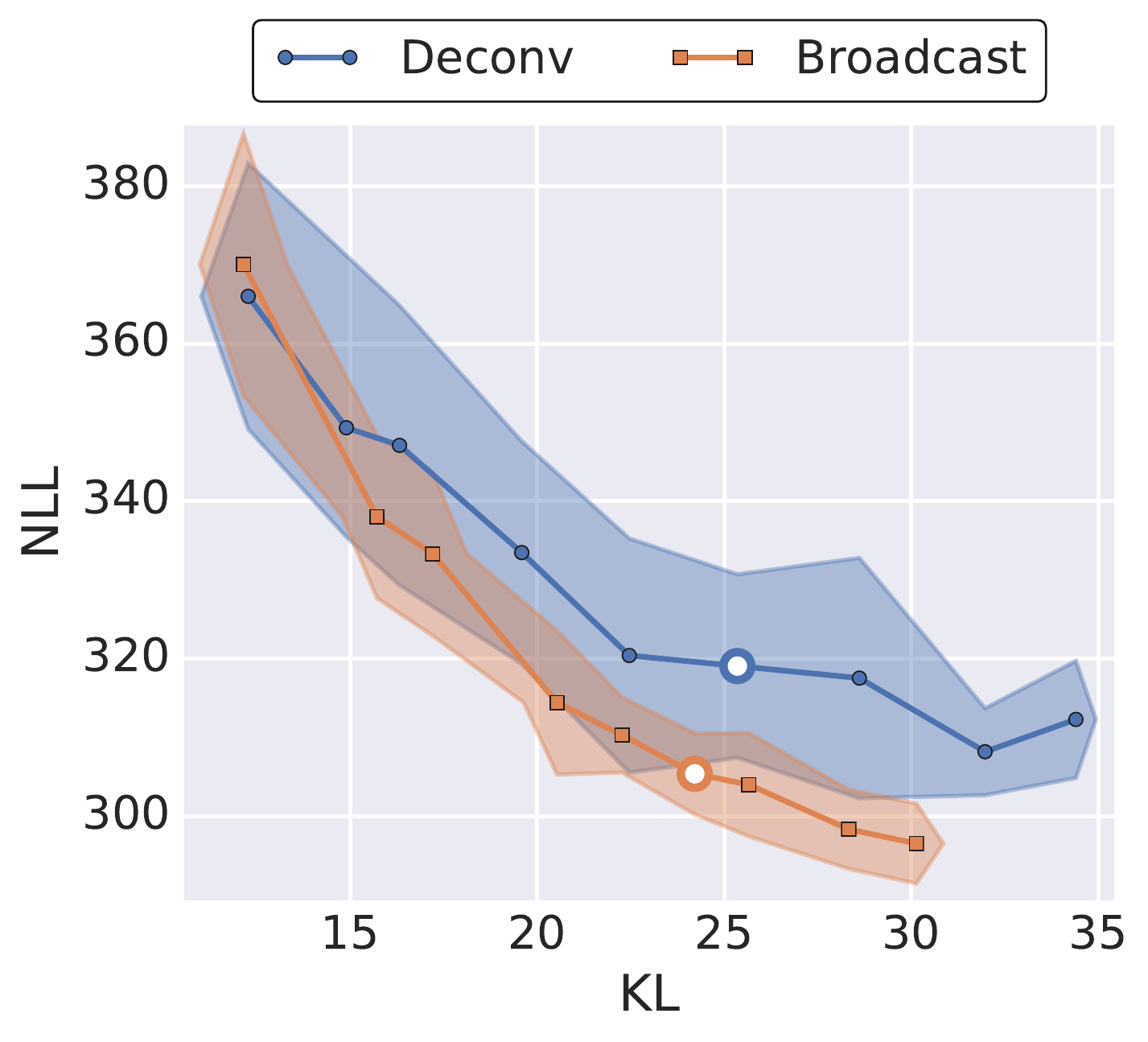}
  \end{subfigure}
  \hfill
  \begin{subfigure}{0.49\textwidth}
  \caption{Reconstruction vs MIG}
  \includegraphics[width=1.0\linewidth]{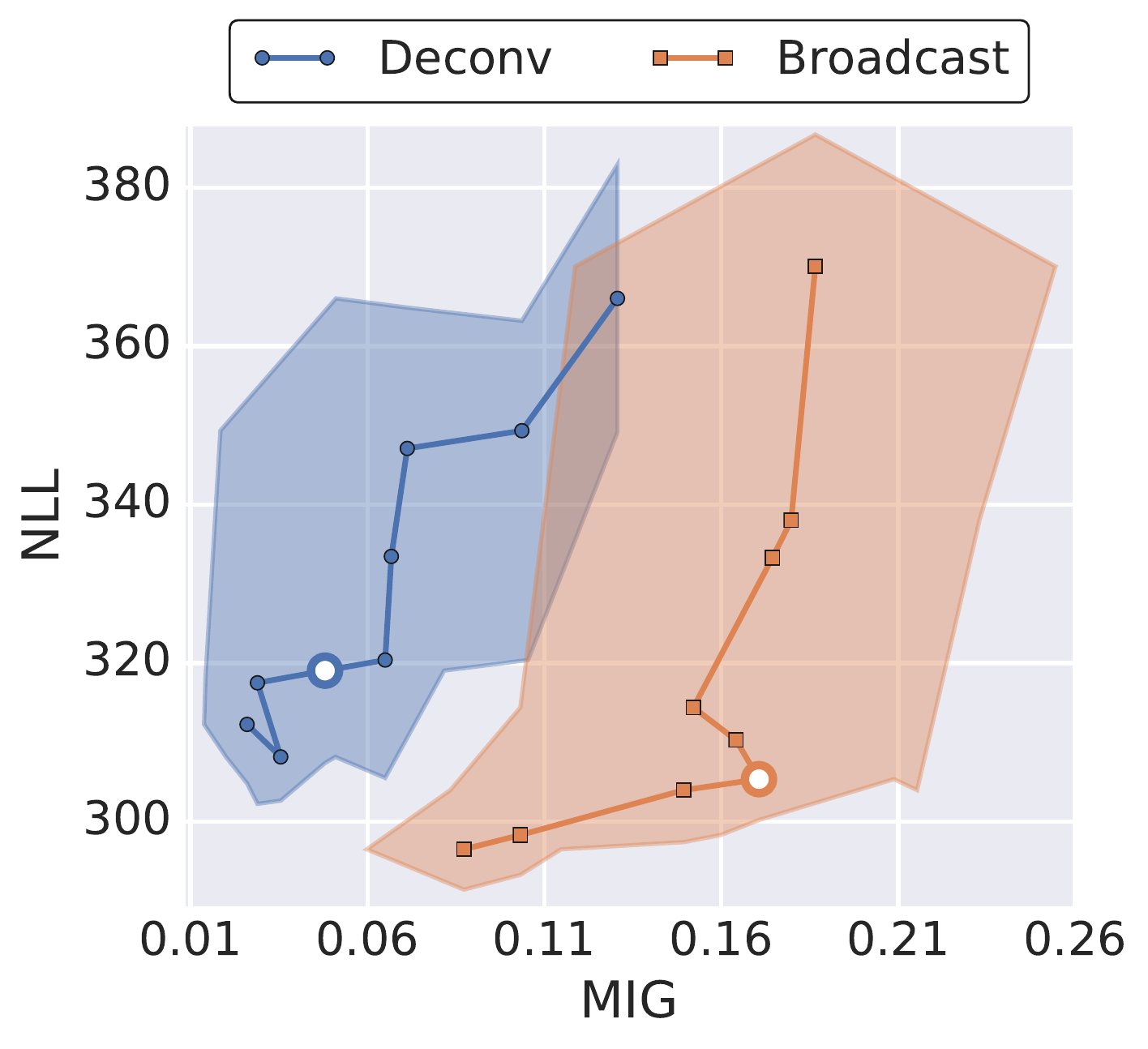}
  \end{subfigure}
  \vspace{-10pt}
  \caption{\textbf{Rate-distortion proxy curves.} We swept $\beta$ log-linearly from 0.4 to 5.4 and for each value trained 10 replicas each of Deconv \betavae (blue) and Spatial Broadcast \betavae (orange) on colored sprites. The dots show the mean over these replicas for each $\beta$, and the shaded region shows the hull of one standard deviation. White dots indicate $\beta=1$.
  \textbf{(a)} Reconstruction (Negative Log-Likelihood, NLL) vs KL. $\beta < 1$ yields low NLL and high KL (bottom-right of figure), whereas $\beta > 1$ yields high NLL and low KL (top-left of figure). See \citet{Alemi2017} for details. Spatial Broadcast \betavae shows a better rate-distortion curve than Deconv \betavae. \textbf{(b)} Reconstruction vs MIG metric. $\beta < 1$ correspond to lower NLL and low MIG regions (bottom-left of figure), and $\beta > 1$ values correspond to high NLL and high MIG scores (towards top-right of figure). Spatial Broadcast \betavae is better disentangled (higher MIG scores) than Deconv \betavae.}\label{fig:rate_distortion}
\end{figure}

Figure \ref{fig:rate_distortion} shows performance of a \betavae with and without the Spatial Broadcast decoder for a range of values of $\beta$. Not only does the Spatial Broadcast decoder improve disentangling, it also yields a lower rate-distortion curve, hence learns to more efficiently represent the data than the same model with a DeConv decoder.

\subsection{Datasets without positional variation}\label{S:no_position}

\begin{figure}[t!]
  \centering
  \hfill
  \begin{subfigure}[b]{0.48\textwidth}
  \includegraphics[width=1.0\linewidth]{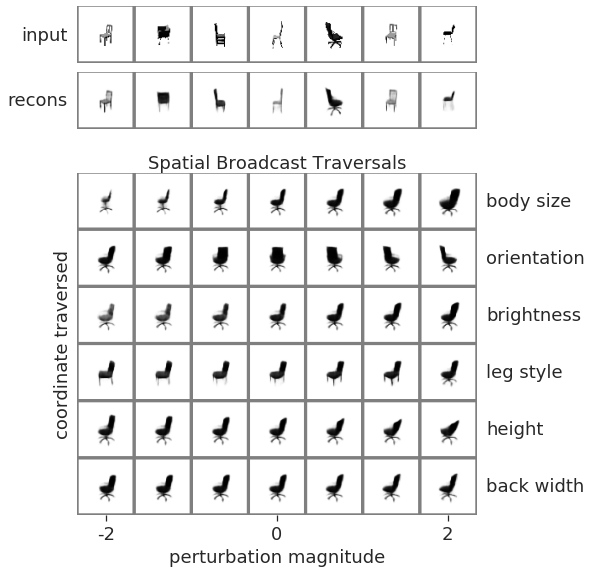}
  \end{subfigure}
  %
  \begin{subfigure}[b]{0.51\textwidth}
  \includegraphics[width=1.0\linewidth]{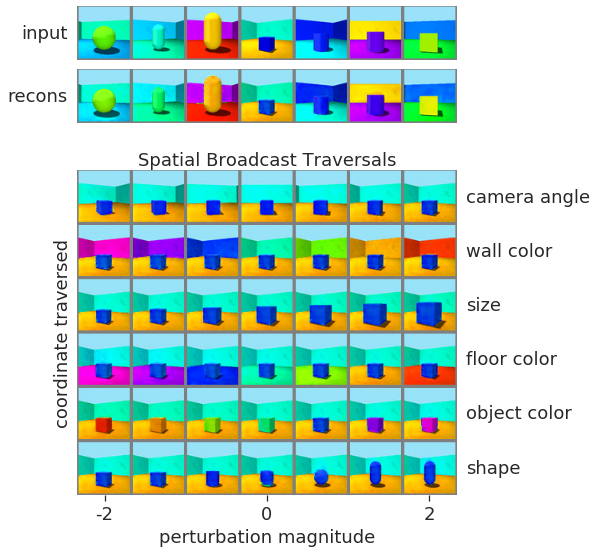}
  \end{subfigure}
  \hfill %
  \vspace{-15pt}
  \caption{\textbf{Traversals for datasets with no positional variation.} A Spatial Broadcast VAE shows good reconstructions and disentangling on the Chairs dataset \citep{Aubry_etal_2014} and the 3D Object-in-Room dataset \citep{Kim_Mnih_2017}. As in Figures \ref{fig:colored_sprites:vae} and \ref{fig:colored_sprites:factorvae}, the models have 10 latent coordinates, though in these traversals the 4 non-coding ones are omitted.}
  \label{fig:chair_objects_room}
  \vspace{-5pt}
\end{figure}

The colored sprites dataset discussed in Section \ref{S:csprites} seems particularly well-suited for the Spatial Broadcast decoder because X- and Y-position are factors of variation.
However, we also evaluate the architecture on datasets that have no positional factors of variation: Chairs and 3D Object-in-Room datasets \citep{Aubry_etal_2014, Kim_Mnih_2017}.
In the latter, the factors of variation are highly non-local, affecting multiple regions spanning nearly the entire image.
We find that on both datasets a Spatial Broadcast VAE learns representations that look very well-disentangled, seemingly as well as SOTA methods on these datasets and without any modification of the standard VAE objective \citep{Kim_Mnih_2017, higgins2017}.
See Figure \ref{fig:chair_objects_room} for results (and supplementary Figure \ref{fig:chair_objects_room_extra} for additional traversals).

While this shows that the Spatial Broadcast decoder does not hurt when used on datasets without positional variation, it may seem unlikely that it would help in this context.
However, we show in Appendix \ref{S:circles_geometry} that it actually can help to some extent on such datasets.
We attribute this to its using a shallower network and no upsampling deconvolutions (which have been observed to cause optimization difficulties in a variety of settings \citep{liu2018, odena2016}).

%% file: small_objects.tex
\subsection{Datasets with small objects}\label{S:small_objects}

\begin{figure}[t!]
  \centering
  \includegraphics[width=0.95\linewidth]{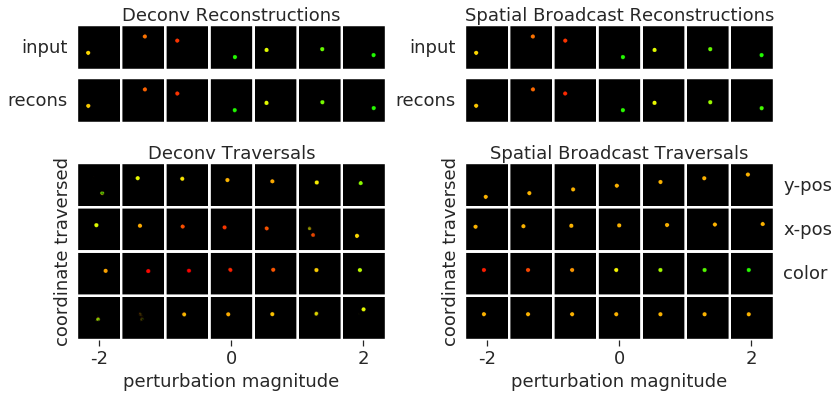}
  \vspace{-5pt}
  \caption{\textbf{Disentangling small objects.} A DeConv VAE \emph{(left)} learns a highly entangled and discontinuous representation of this dataset of small hue-varying circles, while a Spatial Broadcast VAE \emph{(right)} disentangles the dataset well. Here traversals of only the most relevant 4 of the models' 10 latent coordinates are shown (the other latent coordinates are non-coding).}
  \label{fig:circles_small}
  \vspace{-10pt}
\end{figure}

In exploring datasets with objects varying in position, we often find a (standard) DeConv VAE learns a representation that is discontinuous with respect to object position.
This effect is amplified as the size of the object decreases.
This makes sense, because the pressure for a VAE to represent position continuously comes from the fact that an object and a position-perturbed version of itself overlap in pixel space (hence it is economical for the VAE to map noise in its latent samples to local translations of an object).
However, as an object's size decreases, this pixel overlap decreases, hence the pressure for a VAE to represent position continuously weakens.

In this small-object regime the Spatial Broadcast decoder's architectural bias towards representing positional variation continuously proves extremely useful.
We see this in Figure \ref{fig:circles_small} and supplementary Figure \ref{fig:circles_tiny}.

%% file: dependent_factors.tex
\newcommand{\latentgeometrytable}[1]{

\begin{tabular}{c c}
    &
    Dataset Distribution \hspace{15pt} Ground Truth Factors \hspace{15pt} Dataset Samples
    \\
    &
    \includegraphics[width=0.24\linewidth]{figures/circles/#1/#1_dist.png}
    \hspace{10pt}
    \includegraphics[width=0.24\linewidth]{figures/circles/#1/#1_grid.png}
    \hspace{5pt}
    \includegraphics[width=0.24\linewidth]{figures/circles/#1/#1_samples.png}
    \\
    & MIG Metric Values \hspace{25pt} \textcolor{BrickRed}{Worst Replica} \hspace{50pt} \textcolor{OliveGreen}{Best Replica}
    \\
    \begin{turn}{90}
        \begin{tabular}{c c}
        \qquad \qquad DeConv
        \end{tabular}
    \end{turn} \hspace{0pt} 
    & \includegraphics[width=0.8\linewidth]{figures/circles/#1/#1_results_deconv.png}
    \\
    \begin{turn}{90}
        \begin{tabular}{c c}
        \qquad Spatial Broadcast
        \end{tabular}
    \end{turn} \hspace{0pt} 
    & \includegraphics[width=0.8\linewidth]{figures/circles/#1/#1_results.png}
\end{tabular}
}

\subsection{Latent space geometry visualization}\label{S:latent_geometry}

\begin{figure}[t!]
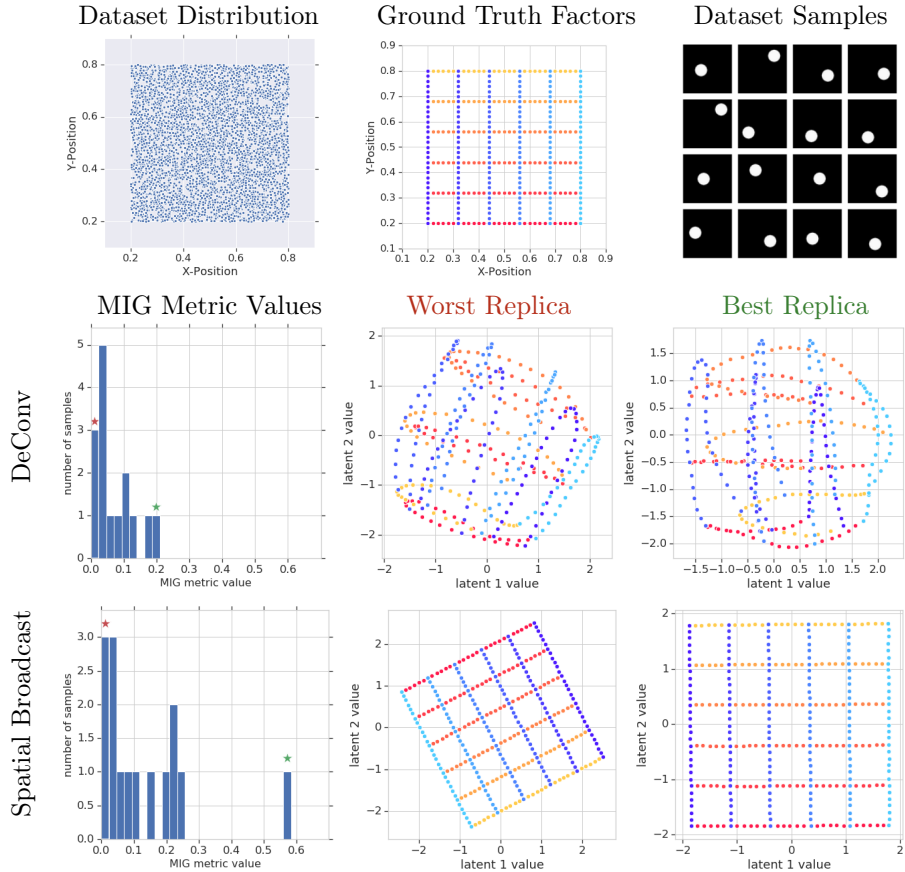

\centering
\resizebox{0.9\textwidth}{!}{
\latentgeometrytable{x_y}
}
\vspace*{-10pt}
\caption{\textbf{Latent space geometry analysis of x-y dataset.}  Comparison of DeConv and Spatial Broadcast decoders in a standard VAE on a dataset of circles varying in position. On the top row we see: \emph{(left)} the data generative factor distribution uniform in X- and Y-position, \emph{(middle)} a grid of points in generative factor space spanning the data distribution, and \emph{(right)} 16 sample images from the dataset. The next two rows show a analysis of the DeConv VAE and Spatial Broadcast VAE on this dataset. In each we see a histogram of MIG metric values over 15 independent replicas and a latent space geometry visualization for the replica with the worst MIG and the replica with the best MIG (highlighted in the MIG histograms by the colored stars). These geometry plots show the embedding of the ground truth factor grid in latent space with respect to the two (out of 10) latent components with the smallest mean variance, i.e. the most informative latent components. The Deconv decoder gives rise to highly entangled latent spaces, while the Spatial Broadcast decoder learns embeddings that are extremely well-disentangled (in fact, nearly linear), even in the worst performing replica. Note that the MIG does not capture this contrast because it is very sensitive to rotation in the latent space.}
\label{fig:latent_space_geometry}
\end{figure}

Evaluating the quality of a representation can be challenging and time-consuming.
While a number of metrics have been proposed to quantify disentangling, many of them have serious shortcomings and there is as yet no consensus in the literature which to use \citep{Locatello_etal_2018}.
We believe it is impossible to quantify how good a representation is with a single scalar, because there is a  fundamental trade-off between how much information a representation contains and how well-structured the representation is.
This has been noted by others in the disentangling literature \citep{Ridgeway_Mozer_2018, Eastwood_Williams_2018}.
This disentangling-distortion trade-off is a recapitulation of the rate-distortion trade-off \citep{Alemi2017} and can be seen first-hand in Figure \ref{fig:rate_distortion}.
We would like representations that both reconstruct well and disentangle well, but exactly how to balance these two factors is a matter of subjective preference (and surely depends on the dataset).
Any scalar disentangling metric will implicitly favor some arbitrary disentangling-reconstruction potential.

In addition to this unavoidable limitation of disentangling metrics, we found that the MIG metric (while perhaps more accurate than other existing metrics) does not capture the intuitive notion of disentangling because:
\begin{itemize}
    \item It depends on a choice of basis for the ground truth factors, and heavily penalizes rotation of the representation with respect to this basis. Yet it is often unclear what the correct basis for the ground truth factors is (e.g. RGB vs HSV vs HSL). For example, see the bottom row of Figure \ref{fig:latent_space_geometry}.
    \item It is invariant to a folding of the representation space, as long as the folds align with the axes of variation. See the middle row of Figure \ref{fig:latent_space_geometry} for an example of a double-fold in the latent space which isn't penalized by the MIG metric.
\end{itemize}

Due to the subjective nature of disentangling and the difficulty in defining appropriate metrics, we put heavy emphasis on latent space visualization as a means for representational analysis.
Latent space traversals have been extensively used in the literature and can be quite revealing \citep{higgins2017, higgins2017scan}.
However, in our experience, traversals suffer two shortcomings:
\begin{itemize}
    \item Some latent space entanglement can be difficult for the eye to perceive in traversals. For example, a slight change in brightness in a latent traversal that represents changing position can easily go unnoticed.
    \item Traversals only represent the latent space geometry around one point in space, and cross-referencing corresponding traversals between multiple points is quite time-consuming.
\end{itemize}

Consequently, we caution the reader against relying too heavily on traversals when evaluating latent space geometry.
We propose an additional method for analyzing latent spaces, which we found very useful in our research.
This is possible when there are known generative factors in the dataset and aims to directly view the embedding of generative factor space in the model's latent space: We plot in latent space the locations corresponding to a grid of points in generative factor space.
While this is can only visualize the latent embedding of a 2-dimensional subspace of generative factor space, it can be very revealing of the latent space geometry.

We showcase this analysis method in Figure \ref{fig:latent_space_geometry} on a dataset of circles varying in X- and Y-position.
(See Appendix \ref{S:circles_geometry} for similar analyses on many more datasets.)
We compare the representations learned by a DeConv VAE and a Spatial Broadcast VAE.
This reveals a stark contrast: The Spatial Broadcast latent geometry is a near-perfect Euclidean transformation, while the Deconv decoder model's representations are very entangled.

The MIG metric does not capture this difference well: It gives a score near zero to a Spatial Broadcast model with a latent space rotated with respect to the generative factors and a greater score to a highly entangled Deconv model.
In practice we care primarily about compositionality of (and hence disentangling of) subspaces of generative factor space and not about axis-alignment within these subspaces.
For example, rotation within X-Y position subspace is acceptable, whereas rotation in position-color space is not.
This naturally poses a great challenge for both designing disentangling metrics and formally defining disentangling \citep[cf.][]{higgins2018}.
We believe that additional structure over the factors of variation can be inferred from temporal correlations, the structure of a reinforcement learning agent's action space, and other effects not necessarily captured by a static-image dataset.
Nonetheless, inductive biases like the Spatial Broadcast decoder can be explored in the context of static images, as they will likely also help representation learning in more complete contexts.

\subsection{Datasets with dependent factors}\label{S:dependent_factors}

\begin{figure}[t!]
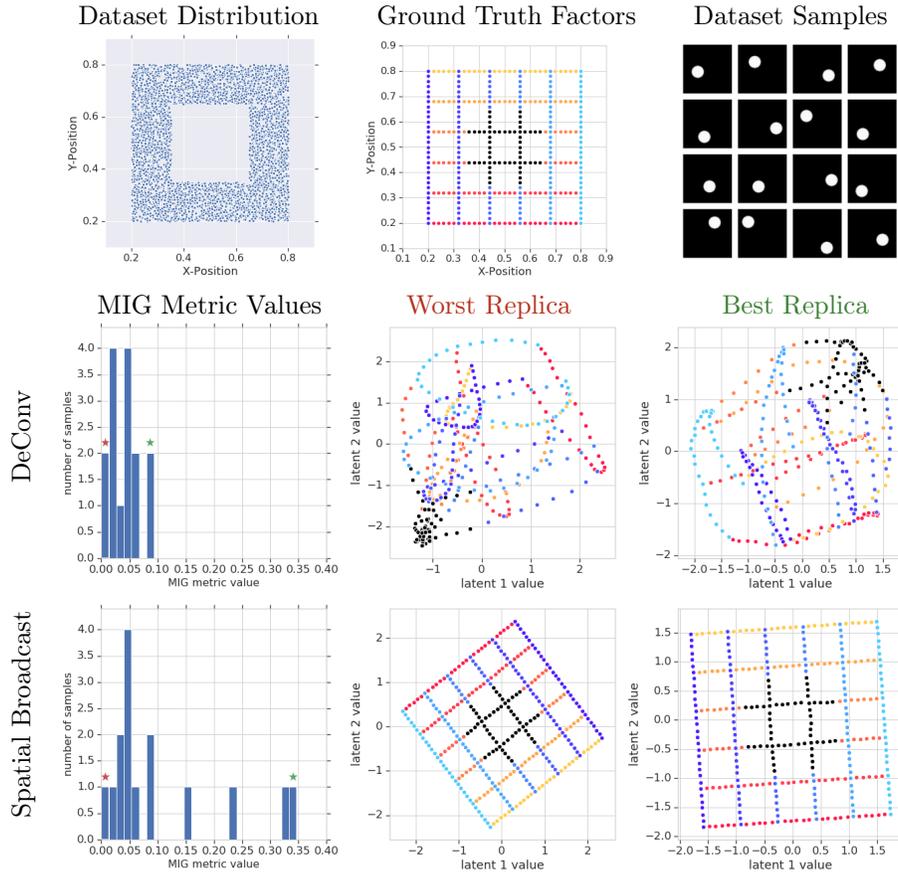

\centering
\resizebox{0.9\textwidth}{!}{
\latentgeometrytable{x_y_center}
}
\vspace*{-10pt}
\caption{\textbf{Latent space geometry analysis of x-y dataset with dependent factors.}
Analogous to Figure \ref{fig:latent_space_geometry}, except the dataset has a large held-out hole in generative factor space (see dataset distribution in top-left), hence the generative factors are not independent. For the latent geometry visualizations, we do evaluate the representation of images in this held-out hole, which are shown as black dots. This tests for generalization, namely extrapolation in pixel space and interpolation in generative factor space. The Spatial Broadcast VAE generalizes extremely well, again with representations looking nearly linear, even through the held-out hole. In contrast, the DeConv VAE learns highly entangled representations. As in Figure \ref{fig:latent_space_geometry}, the MIG does not adequately reflect this difference because of its sensitivity to rotation.}
\label{fig:dependent_factors}
\end{figure}

Our primary motivation for unsupervised representation learning is the prospect of improved generalization and transfer, as discussed in Section \ref{S:intro}.
Consequently, we would like a model to learn compositional representations that can generalize to new feature combinations.
We explore this in a controlled setting by holding out a region of generative factor space from the training dataset, then evaluating the representation in this held-out region.
Figure~\ref{fig:dependent_factors} shows results to this effect, comparing the Spatial Broadcast VAE to a DeConv VAE on a dataset of circles with a held-out region in the middle of X-Y position-space (so the model sees no circles in the middle of the image, indicated by the black dots in the second column).
The Spatial Broadcast VAE generalizes almost perfectly in this case, appearing unaffected by the fact that the data generative factors are no longer independent and extrapolating nearly linearly throughout the held-out region.
In contrast, the DeConv VAE's latent space is highly entangled, even more so than in the case with independent factors of Figure \ref{fig:latent_space_geometry}.
See Appendix \ref{S:circles_geometry} for more analyses of generalization on other datasets.

From the perspective of density modeling, disentangling of the Spatial Broadcast VAE may seem undesirable in this case because it allocates high probability in latent space to a region of low (here, zero) probability in the dataset.
However, from the perspective of representation learning and compositional features, such generalization is a highly desirable property.

%% file: conclusion.tex
\section{Conclusion}\label{S:conclusion}

Here we present and analyze the Spatial Broadcast decoder in the context of Variational Autoencoders.
We demonstrate that it improves learned latent representations, most dramatically for datasets with objects varying in position.
It also improves generalization in latent space and can be incorporated into SOTA models to boost their performance in terms of both disentangling and reconstruction accuracy.
We rigorously analyze the contribution of the Spatial Broadcast decoder on a wide variety of datasets and models using a range of visualizations and metrics.
We hope that this analysis has provided the reader with an intuitive understanding of how and why it improves learned representations.

We believe that learning compositional representations is an important ingredient for flexibility and generalization in many contexts, from supervised learning to reinforcement learning, and the Spatial Broadcast decoder is one step towards robust compositional visual representation learning.

%% file: supplementary.tex
\begin{appendices}\label{S:supp}

\appendix

\input{experiment_details.tex}

\input{ablation.tex}

\input{tiny_objects.tex}

\input{coordconv_vae.tex}

\input{extra_traversals.tex}

\input{hyperparams.tex}

\input{circles_geometry.tex}

\end{appendices}

%% file: experiment_details.tex
\section{Experiment Details}\label{S:experiment_details}

For all VAE models we used a Bernoulli decoder distribution, parametrized by its logits.
It is with respect to this distribution that the reconstruction error (negative log likelihood) was computed.
This could accomodate our datasets, since they were normalized to have pixel values in $[0, 1]$.
We also explored using a Gaussian distribution with fixed variance (for which the NLL is equivalent to scaled MSE), and found that this produces qualitatively similar results and in fact improves stability.
Hence while a Bernoulli distribution usually works, we suggest the reader wishing to experiment with these models starts with a Gaussian decoder distribution with mean parameterized by the decoder network output and variance constant at $0.3$.

In all networks we used ReLU activations, weights initialized by a truncated normal (see \citep{ioffe_2015}), and biases initialized to zero.
We use no other neural network tricks (no BatchNorm or dropout), and all models were trained with the Adam optimizer \citep{Kingma_Ba_2014}.
See below for learning rate details.

\subsection{VAE Hyperparameters}\label{S:experiment_details:vanilla_vae}

For all VAE models except \betavae (shown only in Figure \ref{fig:rate_distortion}), we use a standard VAE loss, namely with a KL term coefficient $\beta = 1$. For FactorVAE we also use $\beta = 1$, as in \citet{Kim_Mnih_2017}.

For the VAE, \betavae, CoordConv VAE and ablation study we used the network parameters in Table \ref{table:network_details}.
We note that, while the Spatial Broadcast decoder uses fewer parameters than the DeConv decoder, it does require about $50\%$ more memory to store the weights.
However, for the 3D Object-in-Room dataset we included three additional deconv layers in the Spatial Droadcast decoder (without these additional layers the decoder was not powerful enough to give good reconstructions on that dataset).

All of these models were trained using a learning rate of $3 \cdot 10^{-4}$ on with batch size 16.
All convolutional and deconvolutional layers have ``same'' padding, i.e. have zero-padded input so that the output shape is $\texttt{input\_shape} \times \texttt{stride}$ in the case of convolution and $\texttt{input\_shape} / \texttt{stride}$ in the case of deconvolution.
\begin{table}[H]
    \centering
    \vspace*{-2pt}
    \resizebox{0.9\textwidth}{!}{
    \begin{minipage}[t]{.37\textwidth}
    \begin{tabular}[t]{c}
    \toprule
    \textbf{\textsc{Encoder}}
    \\
    \midrule
    \texttt{FC($2\times 10$) Output LogNormal}
    \\
    \hline
    \texttt{FC($256$)}
    \\
    \hline
    \texttt{Conv(k=4, s=2, c=64)}
    \\
    \hline
    \texttt{Conv(k=4, s=2, c=64)}
    \\
    \hline
    \texttt{Conv(k=4, s=2, c=64)}
    \\
    \hline
    \texttt{Conv(k=4, s=2, c=64)}
    \\
    \hline
    \texttt{Input Image [$64 \times 64\times C$]}
    \\
    \bottomrule
    \end{tabular}
    \end{minipage}
    \begin{minipage}[t]{.32\textwidth}
    \begin{tabular}[t]{c}
    \toprule
    \textbf{\textsc{DeConv Decoder}}
    \\
    \midrule
    \texttt{Output Logits}
    \\
    \hline
    \texttt{DeConv(k=4, s=2, c=$C$)}
    \\
    \hline
    \texttt{DeConv(k=4, s=2, c=64)}
    \\
    \hline
    \texttt{DeConv(k=4, s=2, c=64)}
    \\
    \hline
    \texttt{DeConv(k=4, s=2, c=64)}
    \\
    \hline
    \texttt{DeConv(k=4, s=2, c=64)}
    \\
    \hline
    \texttt{reshape($2 \times 2 \times 64$)}
    \\
    \hline
    \texttt{FC(256)}
    \\
    \hline
    \texttt{Input Vector [$10$]}
    \\
    \bottomrule
    \end{tabular}
    \end{minipage}
    \begin{minipage}[t]{.27\textwidth}
    \begin{tabular}[t]{c}
    \toprule
    \textbf{\textsc{Broadcast Decoder}}
    \\
    \midrule
    \texttt{Output Logits}
    \\
    \hline
    \texttt{Conv(k=4, s=1, c=$C$)}
    \\
    \hline
    \texttt{Conv(k=4, s=1, c=64)}
    \\
    \hline
    \texttt{Conv(k=4, s=1, c=64)}
    \\
    \hline
    \texttt{append coord channels}
    \\
    \hline
    \texttt{tile($64 \times 64 \times 10$)}
    \\
    \hline
    \texttt{Input Vector [$10$]}
    \\
    \bottomrule
    \end{tabular}
    \end{minipage}
    }
    \vspace{-5pt}
    \caption{Network architectures for Vanilla VAE, \betavae, CoordConv VAE and ablation study. The numbers of layers were selected to minimize the ELBO loss of a VAE on the colored sprites data (see Appendix \ref{S:hyperparams}). Note that for 3D Object-in-Room, we include three additional convolutional layers in the spatial broadcast decoder. Here $C$ refers to the number of channels of the input image, either 1 or 3 depending on whether the dataset has color.}
    \label{table:network_details}
    \vspace{-20pt}
\end{table}

\subsection{FactorVAE}\label{S:experiment_details:factor_vae}

For the FactorVAE model, we used the hyperparameters described in the FactorVAE paper \citep{Kim_Mnih_2017}.
Those network parameters are reiterated in Table \ref{table:network_details_fvae}.
Note that the Spatial Broadcast parameters are the same as for the other models in Table \ref{table:network_details}.
For the optimization hyperparameters we used $\gamma = 35$, a learning rate of $10^{-4}$ for the VAE updates, a learning rate of $2 \cdot 10^{-5}$ for the discriminator updates, and batch size 32.
These parameters generally gave stable results.

However, when training the FactorVAE model on colored sprites we encountered instability during training.
We subsequently did a number of hyperparameter sweeps attempting to improve stability, but to no avail. Ultimately, we used the hyperparameters in Table \ref{table:network_details_fvae}, though even with limited training steps (see Appendix Section \ref{S:experiment_details:steps}) about $15\%$ of seeds diverged before training completed for both Spatial Broadcast and Deconv decoder.

\begin{table}[H]
    \begin{minipage}[t]{.37\textwidth}
    \begin{tabular}[t]{c}
    \toprule
    \textsc{\textsc{Encoder}}
    \\
    \midrule
    \texttt{FC($2\times 10$) Output LogNormal}
    \\
    \hline
    \texttt{FC($256$)}
    \\
    \hline
    \texttt{Conv(k=4, s=2, c=64)}
    \\
    \hline
    \texttt{Conv(k=4, s=2, c=64)}
    \\
    \hline
    \texttt{Conv(k=4, s=2, c=32)}
    \\
    \hline
    \texttt{Conv(k=4, s=2, c=32)}
    \\
    \hline
    \texttt{Input Image [$64 \times 64\times C$]}
    \\
    \bottomrule
    \end{tabular}
    \end{minipage}
    \begin{minipage}[t]{.32\textwidth}
    \begin{tabular}[t]{c}
    \toprule
    \textsc{\textsc{DeConv Decoder}}
    \\
    \midrule
    \texttt{Output Logits}
    \\
    \hline
    \texttt{DeConv(k=4, s=2, c=$C$)}
    \\
    \hline
    \texttt{DeConv(k=4, s=2, c=32)}
    \\
    \hline
    \texttt{DeConv(k=4, s=2, c=32)}
    \\
    \hline
    \texttt{DeConv(k=4, s=2, c=64)}
    \\
    \hline
    \texttt{DeConv(k=4, s=2, c=64)}
    \\
    \hline
    \texttt{reshape($2 \times 2 \times 64$)}
    \\
    \hline
    \texttt{FC(256)}
    \\
    \hline
    \texttt{Input Vector [$10$]}
    \\
    \bottomrule
    \end{tabular}
    \end{minipage}
    \begin{minipage}[t]{.25\textwidth}
    \begin{tabular}[t]{c}
    \toprule
    \textsc{\textsc{Broadcast Decoder}}
    \\
    \midrule
    \texttt{Output Logits}
    \\
    \hline
    \texttt{Conv(k=4, s=1, c=$C$)}
    \\
    \hline
    \texttt{Conv(k=4, s=1, c=64)}
    \\
    \hline
    \texttt{Conv(k=4, s=1, c=64)}
    \\
    \hline
    \texttt{append coord channels}
    \\
    \hline
    \texttt{tile($64 \times 64 \times 10$}
    \\
    \hline
    \texttt{Input Vector [$10$]}
    \\
    \bottomrule
    \end{tabular}
    \end{minipage}
    \caption{Network architectures for FactorVAE. The encoder and DeConv decoder architectures are as described in the FactorVAE paper \citep{Kim_Mnih_2017}, and the Spatial Broadcast decoder architecture is the same as for the other models (Table \ref{table:network_details}.)}
    \label{table:network_details_fvae}
\end{table}

\subsection{Datasets}\label{S:experiment_details:datasets}

All datasets were rendered in images of size $64\times 64$ and normalized to $[0, 1]$.

\textbf{Colored Sprites:}

For this dataset, we use the binary dsprites dataset open-sourced in \citep{dsprites17}, but multiplied by colors sampled in HSV space uniformly within the region $H \in [0.0, 1.0]$, $S \in [0.3, 0.7]$, $V \in [0.3, 0.7]$.
Sans color, there are 737,280 images in this dataset.
However, we sample the colors online from a continuous distribution, effectively making the dataset size infinite.

\textbf{Chairs:}

This dataset is open-sourced in \citep{Aubry_etal_2014}. This dataset, unlike all others we use, has only a single channel in its images.
It contains 86,366 images.

\textbf{3D Object-in-Room:}

This dataset was used extensively in the FactorVAE paper \citep{Kim_Mnih_2017}.
It consists of an object in a room and has 6 factors of variation: Camera angle, object size, object shape, object color, wall color, and floor color.
The colors are sampled uniformly from a continuous set of hues in the range $[0.0, 0.9]$.
This dataset contains 480,000 images, procedurally generated as all combinations of 10 floor hues, 10 wall hues, 10 object hues, 8 object sizes, 4 object shapes, and 15 camera angles.

\textbf{Circles Datasets:}

To more thoroughly explore datasets with a variety of distributions, factors of variation, and held-out test sets we created our own datasets using the Spriteworld environment (\citet{spriteworld}, available from \url{https://github.com/deepmind/spriteworld/}).
We did not use the Reinforcement Learning aspects of Spriteworld, but used Spriteworld's factor distribution library and renderer, so we created ``tasks'' with maximum episode length 1 (so the environment resets every step) and used a dummy agent to record image observations.
All images were rendered with an anti-aliasing factor of 5.
For the results in this paper, we used an earlier version of Spriteworld which used a PyGame renderer instead of the PIL renderer currently in the library, but this choice of renderer does not affect our results.
We used this Spriteworld-based dataset generator for the datasets in Section \ref{S:latent_geometry}.
In these datasets we control subsets of the following factors of variation: X-position, Y-position, Size, Color.
We generated five datasets in this way, which we call \texttt{X-Y}, \texttt{X-H}, \texttt{R-G}, \texttt{X-Y-H Small}, and \texttt{X-Y-H Tiny}, and can be seen in Figures 
(Fig~\ref{fig:circles:x_y}), (Fig~\ref{fig:circles:x_h}), (Fig~\ref{fig:circles:r_g}), (Fig~\ref{fig:circles_small}), and (Fig~\ref{fig:circles_tiny}) respectively.

Table \ref{table:circles_details} shows the values of these factors for each dataset.
Note that for some datasets we define the color distribution in RGB space, and for others we define it in HSV space.

To create the datasets with dependent factors, we hold out one quarter of the dataset (the intersection of half of the ranges of each of the two factors), either centered within the data distribution or in the corner.

For each dataset we generate 500,000 randomly sampled training images.

\begin{table}[h!]
    \begin{center}
    \begin{tabular}{c V{3} c|c|c|c|c|}
    & \texttt{X} & \texttt{Y} & \texttt{Size} & \texttt{(H, S, V)} & \texttt{(R, G, B)}
    \\
    \hlineB{3}
    \texttt{X-Y}
    & [0.2, 0.8] & [0.2, 0.8] & 0.2 & N/A & (1.0, 1.0, 1.0)
    \\
    \hline
    \texttt{X-H}
    & [0.2, 0.8] & 0.5 & 0.3 & ([0.2, 0.8], 1.0, 1.0) & N/A
    \\
    \hline
    \texttt{R-G}
    & 0.5 & 0.5 & 0.5 & N/A & ([0.4, 0.8], [0.4, 0.8], 1.0)
    \\
    \hline
    \texttt{X-Y-H Small}
    & [0.2, 0.8] & [0.2, 0.8] & 0.1 & ([0.2, 0.8], 1.0, 1.0) & N/A
    \\
    \hline
    \texttt{X-Y-H Tiny}
    & [0.2, 0.8] & [0.2, 0.8] & 0.075 & ([0.2, 0.8], 1.0, 1.0) & N/A
    \\
    \hline
    \end{tabular}
    \caption{Circles datasets details. Each row represents a different dataset, and each column shows a generative factor's range for the datasets.}
    \label{table:circles_details}
    \end{center}
\end{table}

\subsection{Training Steps}\label{S:experiment_details:steps}

The number of training steps for each model on each dataset can be found in Table \ref{table:steps}.
In general, for each dataset we used enough training steps so that all models converged.
Note that while the training iterations is different for FactorVAE  than for the other models on colored sprites (due to instability of FactorVAE), this has no bearing on our results because we do not compare across models.
We only compare across decoder architectures, and we always used the same training steps for both DeConv and Spatial Broadcast decoders within each model.

\begin{table}[h!]
    \begin{center}
    \begin{tabular}{c V{3} c | c |}
    & VAE & FactorVAE
    \\
    \hlineB{3}
    \textsc{Colored Sprites}
    & $1.5 \cdot 10^6$ & $1 \cdot 10^6$
    \\
    \hline
    \textsc{Chairs}
    & $7 \cdot 10^5$ & N/A
    \\
    \hline
    \textsc{3D Objects}
    & $2.5 \cdot 10^6$ & N/A
    \\
    \hline
    \textsc{Circles}
    & $5 \cdot 10^5$ & $5 \cdot 10^5$
    \\
    \hline
    \end{tabular}
    \caption{Number of training steps for each model on each dataset. Here the VAE column includes \betavae, Deconv VAE, Spatial Broadcast VAE, CoordConv VAE, and the ablation study.}
    \label{table:steps}
    \end{center}
\end{table}

%% file: ablation.tex
\section{Ablation Study}\label{S:ablation}

One aspect of the Spatial Broadcast decoder is the concatenation of constant coordinate channels to its tiled input latent vector.
While our justification of its performance emphasizes the simplicity of computation it affords, it may seem possible that the coordinate channels are only used to provide positional information and the simplicity of this positional information (linear meshgrid) is irrelevant.
Here we perform an ablation study to demonstrate that this is not the case; the organization of the coordinate channels is important.
For this experiment, we randomly permute the coordinate channels through space.
Specifically, we take the $[h \times w \times 2]$-shape coordinate channels and randomly permute the $h \cdot w$ entries.
We keep each $(i, j)$ pair together to ensure that after the shuffling each location does still have a unique pair of coordinates in the coordinate channels.
Importantly, we only shuffle the coordinate channels once, then keep them constant throughout training.

\begin{figure}[t!]
  \centering
  \includegraphics[width=0.85\linewidth]{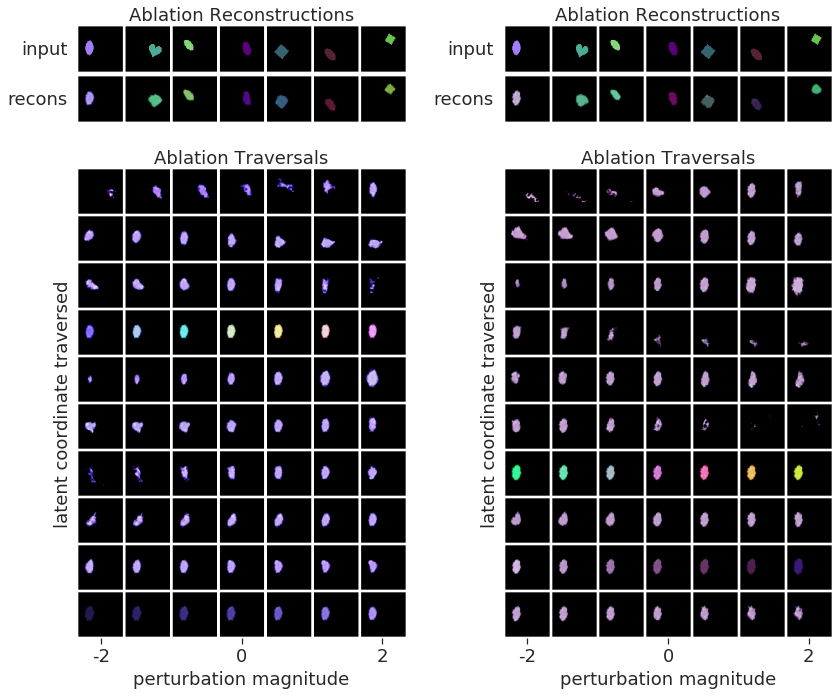}
  \caption{\textbf{Ablation Study} Here we see traversals from a Spatial Broadcast VAE with a random (though constant throughout training) shuffling of the coordinate channels. Different models (with different shufflings) are show on left and the right. We see that the this shuffling negatively impacts both reconstructions and traversals, hence the linear organization of the coordinate channels is important for the model's performance. Quantitatively, this shuffling causes the MIG to drop from $0.147$ to $0.104 \pm 0.025$ (though still better than a DeConv VAE's $0.052$) and the ELBO loss to drop from $329$ to $362.8 \pm 2.90$ (worse than the DeConv VAE's $347$). The shuffled-coordinate model has KL loss $26.56 \pm 0.47$, number of relevant latents $8.99 \pm 0.37$ and negative log likelihood $336.3 \pm 2.98$.}
  \label{fig:ablation}
  \vspace*{-1em}
\end{figure}

Figure \ref{fig:ablation} shows reconstructions and traversals for two replicas (with different shuffled coordinate channels).
Both disentangling and reconstruction accuracy are significantly reduced.

%% file: tiny_objects.tex
\section{Disentangling Tiny Objects}\label{S:tiny_objects}

\begin{figure}[h]
  \centering
  \includegraphics[width=1.0\linewidth]{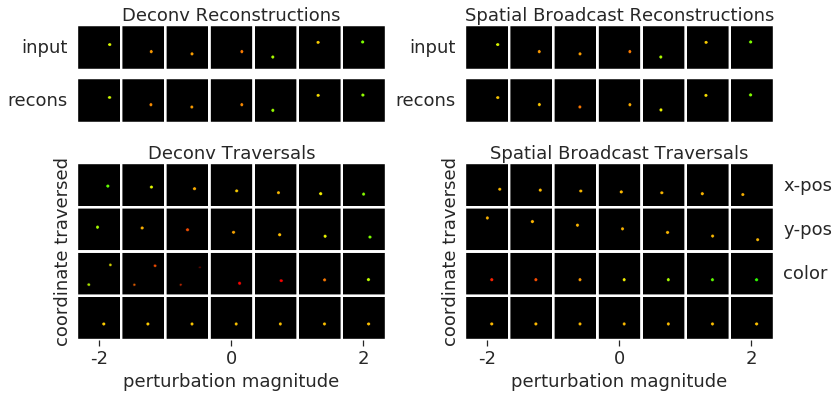}
  \caption{\textbf{Disentangling tiny objects.} A DeConv VAE \emph{(left)} learns a highly entangled and discontinuous representation of this dataset of tiny hue-varying circles. In contrast, a Spatial Broadcast VAE \emph{(right)} disentangles this dataset. The circles are $0.075$ times the frame-width, while those in Figure \ref{fig:circles_small} are $0.1$ times the frame-width.}
  \label{fig:circles_tiny}
\end{figure}

In the dataset of small colored circles shown in Figure \ref{fig:circles_small} the circle diameter is $0.1$ times the frame-width.
We also generated a dataset with circles $0.075$ times the frame-width, and Figure \ref{fig:circles_tiny} shows similar results on this dataset (albeit more difficult for the eye to make out).
We were surprised to see disentangling of such tiny objects and have not explored the lower object size limit for disentangling with the Spatial Broadcast decoder.

%% file: coordconv_vae.tex
\section{CoordConv VAE}\label{S:coordconv_vae}

CoordConv VAE \citep{liu2018} has been proposed as a decoder architecture to improve the continuity of VAE representations.
CoordConv VAE appends coordinate channels to every feature layer of the standard deconvolutional decoder, yet does not spatially tile the latent vector, hence retains upsampling deconvolutions.

\begin{figure}[h]
  \centering
  \begin{subfigure}[b]{0.35\textwidth}
  \includegraphics[width=1.0\linewidth]{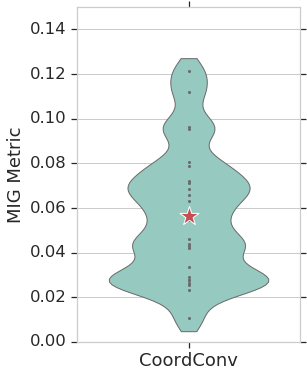}
  \end{subfigure}
  \begin{subfigure}[b]{0.375\textwidth}
  \includegraphics[width=1.0\linewidth]{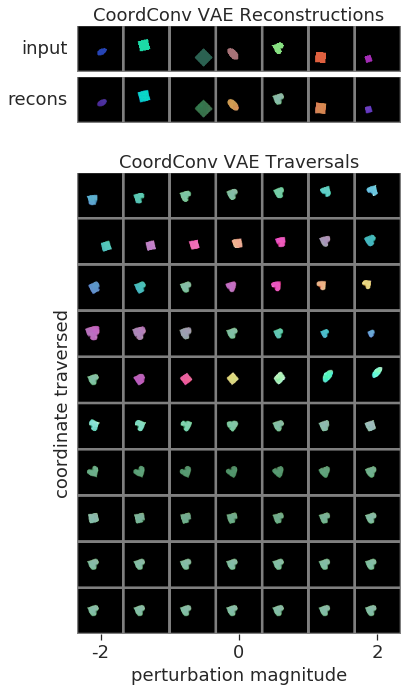}
  \end{subfigure}
  \caption{\textbf{CoordConv VAE on colored sprites.} \emph{(left)} MIG results. Comparing to Figure \ref{fig:colored_sprites:vae}, we see that this achieves far lower scores than a Spatial Broadcast VAE, though about the same as (or maybe slightly better than) a DeConv VAE. \emph{(right)} Latent space traversals are entangled. Note that in contrast to traversal plots in the main text, we show the effect of traversing all 10 latent components (sorted by smallest to largest mean variance), including the non-coding ones (in the bottom rows).}
  \label{fig:colored_sprites:coordconv_vae}
\end{figure}

Figure \ref{fig:colored_sprites:coordconv_vae} shows analysis of this model on the colored sprites dataset.
While the latent space does appear to be continuous with respect to object position, it is quite entangled (far more so than a Spatial Broadcast VAE).
This is not very surprising, since CoordConv VAE uses upscale deconvolutions to go all they way from spatial shape $1\times 1$ to spatial shape $64\times 64$, while in Table \ref{table:upscale_broadcast} we see that introducing upscaling hurts disentangling in a Spatial Broadcast VAE.

%% file: extra_traversals.tex
\section{Extra traversals for datasets without positional variation}\label{S:extra_traversals}

\begin{figure}[h]
  \centering
  \hfill
  \begin{subfigure}[b]{0.48\textwidth}
  \includegraphics[width=1.0\linewidth]{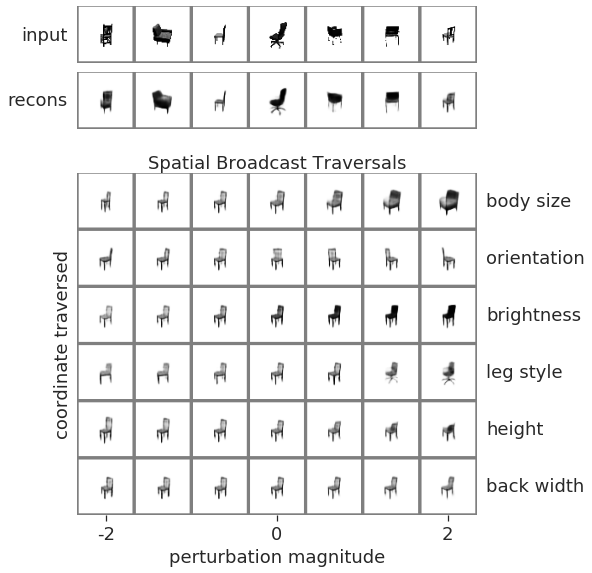}
  \end{subfigure}
  \begin{subfigure}[b]{0.51\textwidth}
  \includegraphics[width=1.0\linewidth]{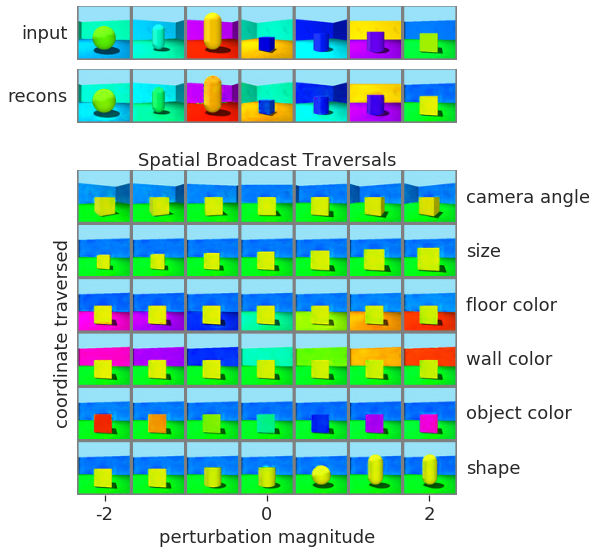}
  \end{subfigure}
  \hfill %
  \caption{\textbf{Traversals for datasets with no positional variation.} Same as Figure \ref{fig:chair_objects_room} but with a different traversal origin in latent space. Cross-referencing with \ref{fig:chair_objects_room}, we see that the Spatial Broadcast VAE does seem to globally disentangle these datasets.}
  \label{fig:chair_objects_room_extra}
\end{figure}

As we acknowledge in the main text, a single latent traversal plot only shows local disentangling at one point in latent space.
Hence to support our claim in Section \ref{S:no_position} that the Spatial Broadcast VAE disentangled the Chairs and 3D objects datasets, we show in Figure \ref{fig:chair_objects_room_extra} traversals about a second seed in each dataset for the same models as in Figure \ref{fig:chair_objects_room}.

%% file: hyperparams.tex
\section{Architecture Hyperparameters}\label{S:hyperparams}

In order to remain objective when selecting model hyperparameters for the Spatial Broadcast and Deconv decoders, we chose hyperparameters based on minimizing the ELBO loss, not considering any information about disentangling.
After finding reasonable encoder hyperparameters, we performed large-scale (25 replicas each) sweeps over a few decoder hyperparameters for both the DeConv and Spatial Broadcast decoder on the colored sprites dataset.
These sweeps are revealing of hyperparameter sensitivity, so we report the following quantities for them:

\begin{itemize}
    \item ELBO. This is the evidence lower bound (total VAE loss). It is the sum of the negative log likelihood (NLL) and KL-divergence.
    \item NLL. This is the negative log likelihood of an image with respect to the model's reconstructed distribution of that image. It is a measure of reconstruction accuracy.
    \item KL. This is the KL divergence of the VAE's latent distribution with its Gaussian prior. It measures how much information is being encoded in the latent space.
    \item Latents Used. This is the mean number of latent coordinates with standard deviation less than $0.5$. Typically, a VAE will have some unused latent coordinates (with standard deviation near $1$) and some used latent coordinates. The threshold $0.5$ is arbitrary, but this quantity does provide a rough idea of how many factors of variation the model may be representing.
    \item MIG. The MIG metric.
    \item Factor VAE. This is the metric described in the FactorVAE paper \citep{Kim_Mnih_2017}. We found this metric to be less consistent than the MIG metric (and equally flawed with respect to rotated coordinates), but it qualitatively agrees with the MIG metric most of the time.
\end{itemize}

\subsection{ConvNet Depth}\label{S:cnn_depth}

Table \ref{table:cnn_depth_broadcast} shows results of sweeping over ConvNet depth in the Spatial Broadcast decoder.
This reveals a consistent trend: As the ConvNet deepens, the model moves towards lower rate/higher distortion.
Consequently, latent space information and reconstruction accuracy drop.
Traversals with deeper nets show the model dropping factors of variation (the dataset has 8 factors of variation).

Table \ref{table:cnn_depth_deconv} shows a noisier but similar trend when increasing DeConvNet depth in the DeConv decoder.

\begin{table}[h!]
    \center
    \resizebox{\textwidth}{!}{
    \begin{tabular}{l@{\hskip 5pt}c c c c c c}
    \toprule
    \textsc{ConvNet} & \textsc{ELBO} & \textsc{NLL} & \textsc{KL} & \textsc{Latents Used} & \textsc{MIG} & \textsc{Factor VAE}
    \\
    \midrule
    \textsc{2-Layer} & 339 ($\pm$2.3) & 312 ($\pm$2.7) & 27.5 ($\pm$0.54) & 8.33 ($\pm$0.37) & 0.076 ($\pm$0.038) & 0.187 ($\pm$0.027)
    \\
    \textsc{3-Layer} & \textbf{329 ($\pm$4.1)} & 305 ($\pm$4.6) & 24.4 ($\pm$0.54) & 7.22 ($\pm$0.34) & 0.147 ($\pm$0.057) & 0.208 ($\pm$0.084)
    \\
    \textsc{4-Layer} & 341 ($\pm$6.8) & 318 ($\pm$7.7) & 22.6 ($\pm$0.97) & 5.93 ($\pm$0.51) & 0.157 ($\pm$0.045) & 0.226 ($\pm$0.046)
    \\
    \textsc{5-Layer} & 340 ($\pm$8.8) & 317 ($\pm$9.7) & 22.7 ($\pm$0.99) & 5.70 ($\pm$0.37) & 0.173 ($\pm$0.059) & 0.218 ($\pm$0.030)
    \\
    \bottomrule
    \end{tabular}
    }
    \caption{\textbf{Effect of ConvNet depth on Spatial Broadcast VAE performance.} These results use the colored sprites dataset. Traversals with deeper nets show the model dropping factors of variation (usually color first, then angle, shape, size, and position in that order). The increasing metric values with deeper ConvNets belies the fact that the model is encoding fewer of the 8 factors of variation in the dataset.}
    \label{table:cnn_depth_broadcast}
\end{table}

\begin{table}[h!]
    \centering
    \resizebox{\textwidth}{!}{
    \begin{tabular}{l@{\hskip 5pt} c c c c c c}
    \toprule
    \textsc{ConvNet} & \textsc{ELBO} & \textsc{NLL} & \textsc{KL} & \textsc{Latents Used} & \textsc{MIG} & \textsc{Factor VAE}
    \\
    \midrule
    \textsc{3-Layer} & 372 ($\pm$8.6) & 346 ($\pm$8.9) & 26.8 ($\pm$0.40) & 9.20 ($\pm$0.04) & 0.031 ($\pm$0.018) & 0.144 ($\pm$0.031)
    \\
    \textsc{4-Layer} & 349 ($\pm$9.4) & 322 ($\pm$10.0) & 27.1 ($\pm$0.88) & 8.90 ($\pm$0.24) & 0.025 ($\pm$0.015) & 0.139 ($\pm$0.009)
    \\
    \textsc{5-Layer} & \textbf{340 ($\pm$9.8)} & 314 ($\pm$10.4) & 26.0 ($\pm$1.00) & 7.95 ($\pm$0.64) & 0.056 ($\pm$0.32) & 0.184 ($\pm$0.053)
    \\
    \textsc{6-Layer} & 349 ($\pm$15.0) & 326 ($\pm$16.1) & 23.3 ($\pm$1.42) & 6.36 ($\pm$0.81) & 0.056 ($\pm$0.019) & 0.199 ($\pm$0.029)
    \\
    \bottomrule
    \end{tabular}
    }
    \caption{\textbf{Effect of ConvNet depth on DeConv VAE performance.} These results use the colored sprites dataset. Similarly to Table \ref{table:cnn_depth_broadcast}, deeper convnets cause the model to represent fewer factors of variation.}
    \label{table:cnn_depth_deconv}
\end{table}

\subsection{MLP Depth}\label{S:mlp_depth}

The Spatial Broadcast decoder as presented in this work is fully convolutional.
It contains no MLP.
However, motivated by the need for more depth on the 3D Object-in-Room dataset, we did explore applying an MLP to the input vector prior to the broadcast operation.
We found that including this MLP had a qualitatively similar effect as increasing the number of convolutional layers on the colored sprited dataset, decreasing latent capacity and giving poorer reconstructions.
These results are shown in Table \ref{table:mlp_depth_broadcast}.

However, on the 3D Object-in-Room dataset adding the MLP did improve the model when using ConvNet depth 3 (the same as for colored sprites).
Results of a sweep over depth of a pre-broadcast MLP are shown in Table \ref{table:mlp_depth_broadcast_room}.
As mentioned in Section \ref{S:no_position}, we were able to achieve the same effect by instead increasing the ConvNet depth to 6, but for those interested in computational efficiency using a pre-broadcast MLP may be a better choice for datasets of this sort.

In the DeConv decoder, increasing the MLP layers again has a broadly similar effect as increasing the ConvNet layers, as shown in Table \ref{table:mlp_depth_deconv}.

\begin{table}[h!]
    \centering
    \resizebox{\textwidth}{!}{
    \begin{tabular}{l@{\hskip 10pt} c c c c c c}
    \toprule
    \textsc{MLP} & \textsc{ELBO} & \textsc{NLL} & \textsc{KL} & \textsc{Latents Used} & \textsc{MIG} & \textsc{Factor VAE}
    \\
    \midrule
    \textsc{0-Layer} & \textbf{329 ($\pm$4)} & 305 ($\pm$5) & 24.5 ($\pm$0.54) & 7.21 ($\pm$0.34) & 0.147 ($\pm$0.057) & 0.208 ($\pm$0.084)
    \\
    \textsc{1-Layer} & 330 ($\pm$6) & 307 ($\pm$6) & 23.9 ($\pm$0.72) & 6.68 ($\pm$0.46) & 0.164 ($\pm$0.043) & 0.200 ($\pm$0.034)
    \\
    \textsc{2-Layer} & 349 ($\pm$15) & 327 ($\pm$17) & 21.5 ($\pm$2.13) & 6.03 ($\pm$0.66) & 0.210 ($\pm$0.048) & 0.232 ($\pm$0.045)
    \\
    \textsc{3-Layer} & 392 ($\pm$23) & 399 ($\pm$114) & 15.7 ($\pm$2.93) & 4.17 ($\pm$1.23) & 0.160 ($\pm$0.034) & 0.275 ($\pm$0.064)
    \\
    \bottomrule
    \end{tabular}
    }
    \caption{\textbf{Effect of a pre-broadcast MLP on the Spatial Broadcast VAE's performance.} These results use the colored sprites dataset, on which an MLP seems to hurt performance (though as noted in the text this is not always the case on the 3D Object-in-Room dataset).}
    \label{table:mlp_depth_broadcast}
\end{table}

\begin{table}[h!]
    \centering
    \resizebox{\textwidth}{!}{
    \begin{tabular}{l@{\hskip 10pt} c c c c c c}
    \toprule
    \textsc{MLP} & \textsc{ELBO} & \textsc{NLL} & \textsc{KL} & \textsc{Latents Used} & \textsc{MIG} & \textsc{Factor VAE}
    \\
    \midrule
    \textsc{1-Layer} & \textbf{347 ($\pm$13)} & 321 ($\pm$14) & 25.8 ($\pm$1.3) & 7.97 ($\pm$0.72) & 0.052 ($\pm$0.020) & 0.174 ($\pm$0.043)
    \\
    \textsc{2-Layer} & 352 ($\pm$17) & 328 ($\pm$18) & 23.7 ($\pm$1.7) & 6.68 ($\pm$0.78) & 0.051 ($\pm$0.024) & 0.196 ($\pm$0.024)
    \\
    \textsc{3-Layer} & 365 ($\pm$19) & 345 ($\pm$21) & 19.7 ($\pm$2.4) & 5.24 ($\pm$0.73) & 0.144 ($\pm$0.062) & 0.243 ($\pm$0.043)
    \\
    \bottomrule
    \end{tabular}
    }
    \caption{\textbf{Effect of MLP depth on the DeConv VAE's performance.} Increasing MLP depth seems to have a broadly similar effect as increasing DeConvNet depth, causing the model to represent fewer factors of variation in the data.}
    \label{table:mlp_depth_deconv}
\end{table}

\begin{table}[h!]
    \centering
    \resizebox{\textwidth}{!}{
    \begin{tabular}{l@{\hskip 10pt} c c c c c c}
    \toprule
    \textsc{MLP} & \textsc{ELBO} & \textsc{NLL} & \textsc{KL} & \textsc{Latents Used} & \textsc{MIG} & \textsc{Factor VAE}
    \\
    \midrule
    \textsc{0-Layer} & 4039 ($\pm$3.4) & 4010 ($\pm$3.3) & 29.3 ($\pm$0.46) & 8.73 ($\pm$0.41) & 0.541 ($\pm$0.091) & 0.931 ($\pm$0.043)
    \\
    \textsc{1-Layer} & 4022 ($\pm$3.7) & 4003 ($\pm$3.7) & 19.3 ($\pm$0.35) & 6.30 ($\pm$0.49) & 0.538 ($\pm$0.105) & 0.946 ($\pm$0.043)
    \\
    \textsc{2-Layer} & 4018 ($\pm$3.3) & 3999 ($\pm$3.3) & 18.5 ($\pm$0.30) & 5.94 ($\pm$0.41) & 0.574 ($\pm$0.096) & 0.978 ($\pm$0.027)
    \\
    \textsc{3-Layer} & 4020 ($\pm$3.0) & 4002 ($\pm$2.9) & 18.3 ($\pm$0.38) & 5.73 ($\pm$0.31) & 0.659 ($\pm$0.123) & 0.979 ($\pm$0.037)
    \\
    \bottomrule
    \end{tabular}
    }
    \caption{\textbf{Effect of a pre-broadcast MLP on the Spatial Broadcast VAE's performance, 3D Object-in-Room dataset.} This table is analogous to Table \ref{table:mlp_depth_broadcast}, except using the 3D Object-in-Room dataset. Here the model seems to over-representation the dataset generative factors (or which there are 6 for this dataset) without a pre-broadcast MLP (top row). However, adding a pre-broadcast MLP with 2 or 3 layers gives rise to accurate reconstructions with the appropriate number of used latents and good disentangling. Adding a pre-broadcast MLP like this is an alternative to increasing the ConvNet depth in the model (shown in Figure \ref{fig:chair_objects_room}).}
    \label{table:mlp_depth_broadcast_room}
\end{table}

\subsection{Decoder Upscale Factor}\label{S:dec_upscale}

We acknowledge that there is a continuum of models between the Spatial Broadcast decoder and the Deconv decoder.
One could interpolate from one to the other by incrementally replacing the convolutional layers in the Spatial Broadcast decoder's network by deconvolutional layers with stride 2 (and simultaneously decreasing the height and width of the tiling operation).
Table \ref{table:upscale_broadcast} shows a few steps of such a progression, where (starting from the bottom) 1, 2, and all 3 of the convolutional layers in the Spatial Broadcast decoder are replaced by a deconvolutional layer with stride 2.
We see that this hurts disentangling without affecting the other metrics, further evidence that upscaling deconvolutional layers are bad for representation learning.

\begin{table}[h!]
    \centering
    \resizebox{\textwidth}{!}{
    \begin{tabular}{l@{\hskip 10pt} c c c c c c}
    \toprule
    \textsc{MLP} & \textsc{ELBO} & \textsc{NLL} & \textsc{KL} & \textsc{Latents Used} & \textsc{MIG} & \textsc{Factor VAE}
    \\
    \midrule
    \textsc{0 Upscales} & 329 ($\pm$4.1) & 305 ($\pm$4.6) & 24.4 ($\pm$0.54) & 7.22 ($\pm$0.34) & 0.147 ($\pm$0.057) & 0.208 ($\pm$0.084)
    \\
    \textsc{1 Upscale} & 327 ($\pm$4.4) & 302 ($\pm$4.9) & 24.4 ($\pm$0.55) & 7.29 ($\pm$0.26) & 0.149 ($\pm$0.048) & 0.194 ($\pm$0.026)
    \\
    \textsc{2 Upscales} & 329 ($\pm$4.3) & 304 ($\pm$4.8) & 24.2 ($\pm$0.60) & 7.14 ($\pm$0.42) & 0.122 ($\pm$0.045) & 0.235 ($\pm$0.070)
    \\
    \textsc{3 Upscales} & 330 ($\pm$2.4) & 305 ($\pm$2.7) & 24.2 ($\pm$0.24) & 7.39 ($\pm$0.08) & 0.110 ($\pm$0.032) & 0.182 ($\pm$0.028)
    \\
    \bottomrule
    \end{tabular}
    }
    \caption{\textbf{Effect of upscale deconvolution on the Spatial Broadcast VAE's performance.} These results use the colored sprites dataset. The columns show the effect of repeatedly replacing the convolutional, stride-1 layers in the decoder by deconvolutional, stride-2 layers (starting at the bottom-most layer). This incrementally reduces performance without affecting the other statistics much, testament to the negative impact of upscaling deconvolutional layer on VAE representations.}
    \label{table:upscale_broadcast}
\end{table}

%% file: circles_geometry.tex
\newcommand{\latentgeometrytablesupp}[1]{
\vspace*{-60pt}
\centering
\begin{adjustbox}{totalheight=\textheight - 5\baselineskip}
\begin{tabular}{c c}
    &
    Dataset Distribution \hspace{15pt} Ground Truth Factors \hspace{15pt} Dataset Samples
    \\
    &
    \includegraphics[width=0.24\linewidth]{figures/circles/#1/#1_dist.png}
    \hspace{10pt}
    \includegraphics[width=0.24\linewidth]{figures/circles/#1/#1_grid.png}
    \hspace{5pt}
    \includegraphics[width=0.22\linewidth]{figures/circles/#1/#1_samples.png}
    \\
    & MIG Metric Values \hspace{25pt} \textcolor{BrickRed}{Worst Replica} \hspace{50pt} \textcolor{OliveGreen}{Best Replica}
    \\
    \begin{turn}{90}
        \begin{tabular}{c c}
        \qquad VAE DeConv
        \end{tabular}
    \end{turn} \hspace{0pt} 
    & \includegraphics[width=0.8\linewidth]{figures/circles/#1/#1_results_deconv.png}
    \\
    \begin{turn}{90}
        \begin{tabular}{c c}
        \qquad VAE \\ \qquad Spatial Broadcast
        \end{tabular}
    \end{turn} \hspace{0pt} 
    & \includegraphics[width=0.8\linewidth]{figures/circles/#1/#1_results.png}
    \\
    \begin{turn}{90}
        \begin{tabular}{c c}
        \qquad FactorVAE \\ \qquad DeConv
        \end{tabular}
    \end{turn} \hspace{0pt} 
    & \includegraphics[width=0.8\linewidth]{figures/circles/#1/#1_results_fvae_deconv.png}
    \\
    \begin{turn}{90}
        \begin{tabular}{c c}
        \quad FactorVAE \\ \quad Spatial Broadcast
        \end{tabular}
    \end{turn} \hspace{0pt} 
    & \includegraphics[width=0.8\linewidth]{figures/circles/#1/#1_results_fvae.png}
\end{tabular}
\end{adjustbox}
}

\section{Latent Space Geometry Analysis for Circle Datasets}\label{S:circles_geometry}

We showed visualization of latent space geometry on the circles datasets in Figures \ref{fig:latent_space_geometry} (with independent factors of variation) and \ref{fig:dependent_factors} (with dependent factors of variation).
These figures showcased the improvement that the Spatial Broadcast decoder lends.
However, we also conducted the same style experiments on many more datasets and on FactorVAE models.
In this section we will present these additional results.

We consider three generative factor pairs: (X-Position, Y-Position), (X-Position, Hue), and (Redness, Greenness).
Broadly, the following figures show that the Spatial Broadcast decoder nearly always helps disentangling.
It helps most dramatically on the most positional variation (X-Position, Y-Position) and least significantly when there is no positional variation (Redness, Greenness).

Note, however, that even with no position variation, the Spatial Broadcast decoder does seem to improve latent space geometry in the generalization experiments (Figures \ref{fig:circles:r_g_center} and \ref{fig:circles:r_g_corner}).
We believe this may be due in part to the fact that the Spatial Broadcast decoder is shallower than the DeConv decoder.

Finally, we explore one completely different dataset with dependent factors: A dataset where half the images have no object (are entirely black).
This we do to simulate conditions like that in a multi-entity VAE such as \citep{nash17} when the dataset has a variable number of entities.
These conditions pose a challenge for disentangling, because the VAE objective will wish to allocate a large (low-KL) region of latent space to representing a blank image when there is a large proportion of blank images in the dataset.
However, we do see a stark improvement by using the Spatial Broadcast decoder in this case.

\begin{figure}[t!]
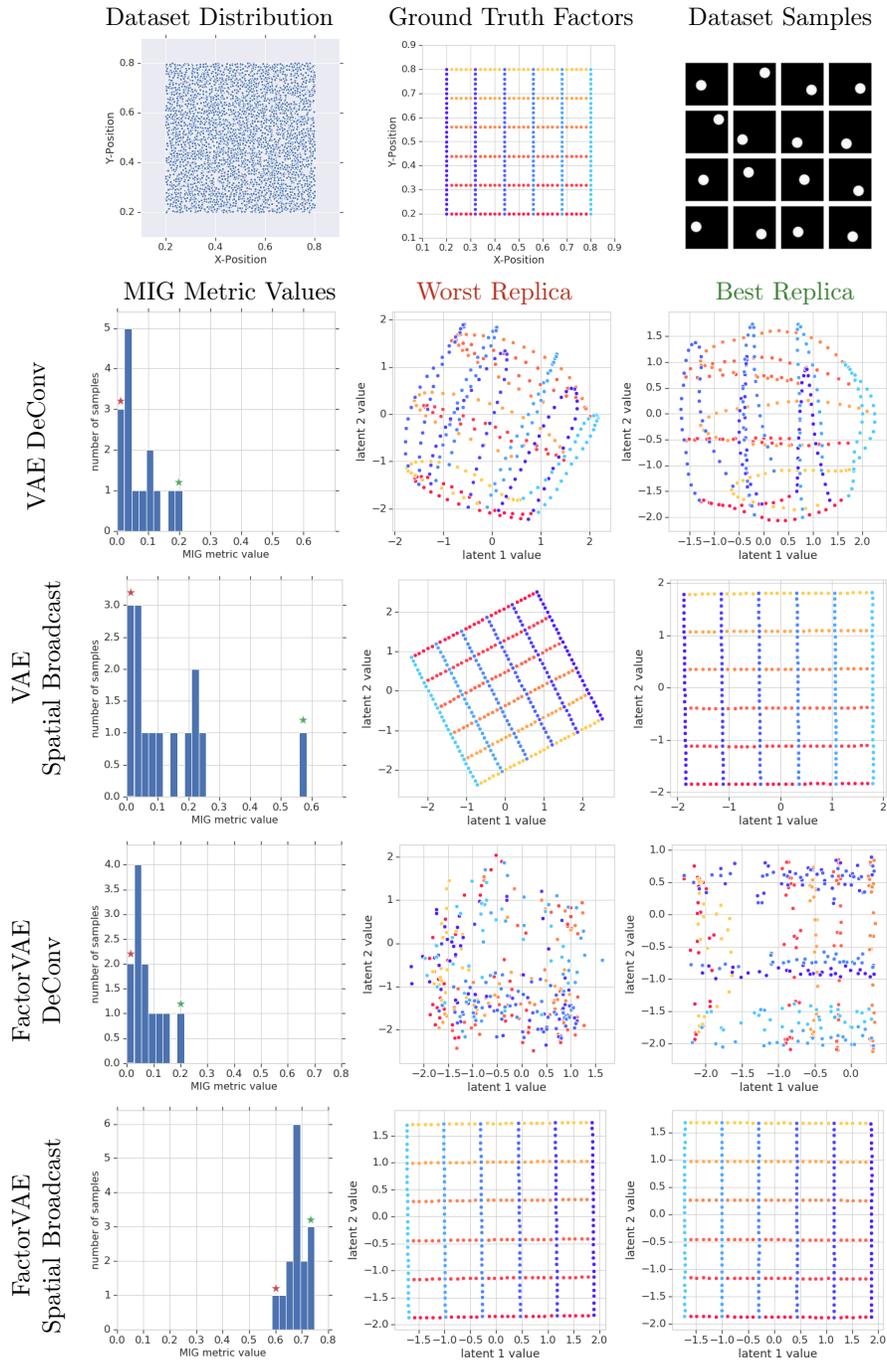

\latentgeometrytablesupp{x_y}
\caption{\textbf{Latent space analysis for independent X-Y dataset.} This is the same as in Figure \ref{fig:latent_space_geometry}, except with the additional FactorVAE results. We see that the Spatial Broadcast decoder improves FactorVAE as well as a VAE (and in the FactorVAE case seems to always be axis-aligned). Note that DeConv FactorVAE has a surprisingly messy latent space --- we found that using a fixed-variance Normal (instead of Bernoulli) decoder distribution improved this significantly, though still not to the level of the Spatial Broadcast FactorVAE. We also noticed in small-scale experiments that including shape or size variability in the dataset helped FactorVAE disentangle as well. However, FactorVAE does seem to be generally quite sensitive to hyperparameters \citep{Locatello_etal_2018}, as adversarial models often are.}
\label{fig:circles:x_y}
\end{figure}

\begin{figure}[t!]
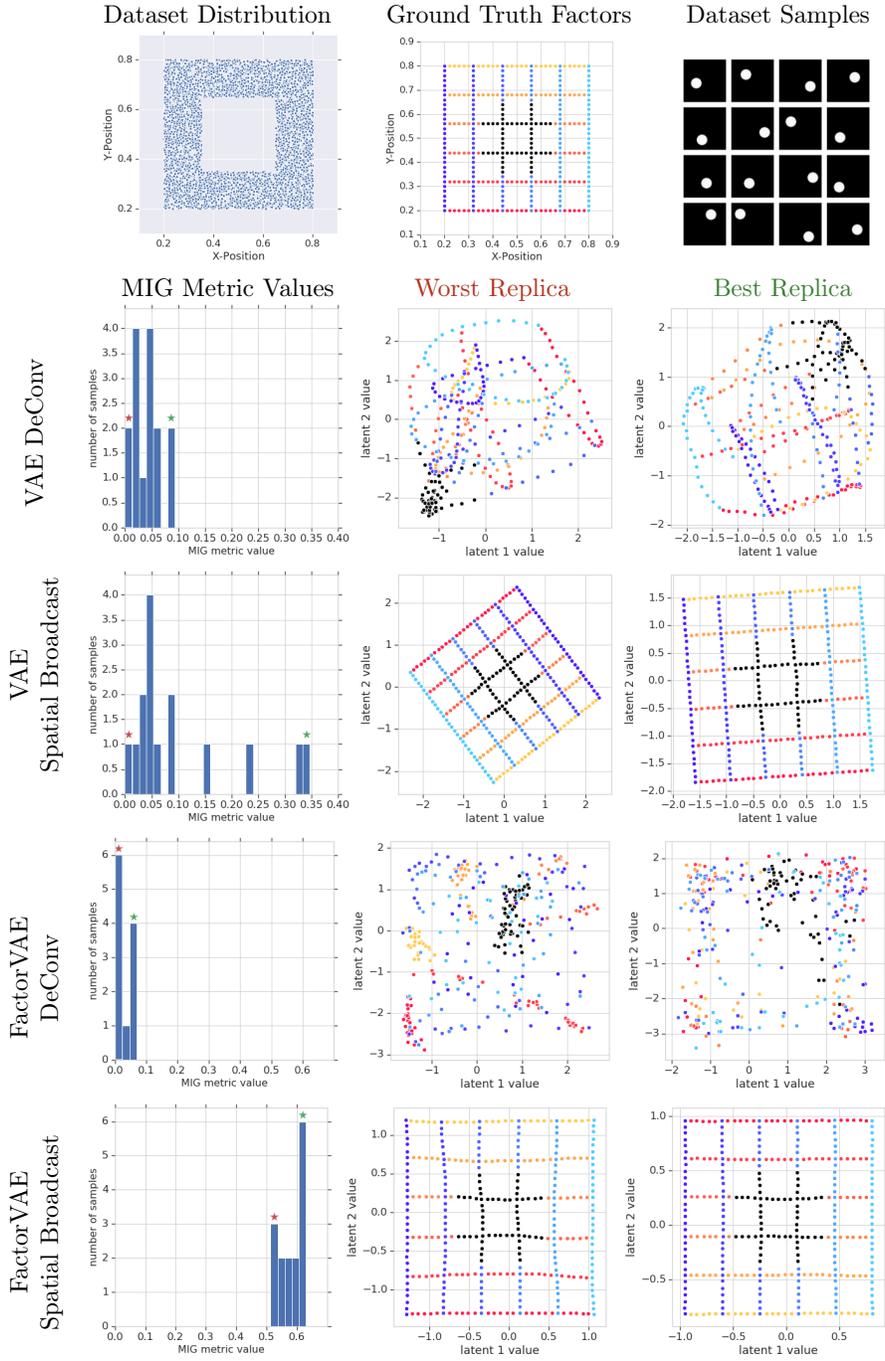

\latentgeometrytablesupp{x_y_center}
\caption{\textbf{Latent space analysis for dependent X-Y dataset.} This is the same as in Figure \ref{fig:dependent_factors}, though with the additional FactorVAE results. Again, the Spatial Broadcast decoder dramatically improves the representation --- its representation looks nearly linear with respect to the ground truth factors, even through the extrapolation region.}
\label{fig:circles:x_y_center}
\end{figure}

\begin{figure}[t!]
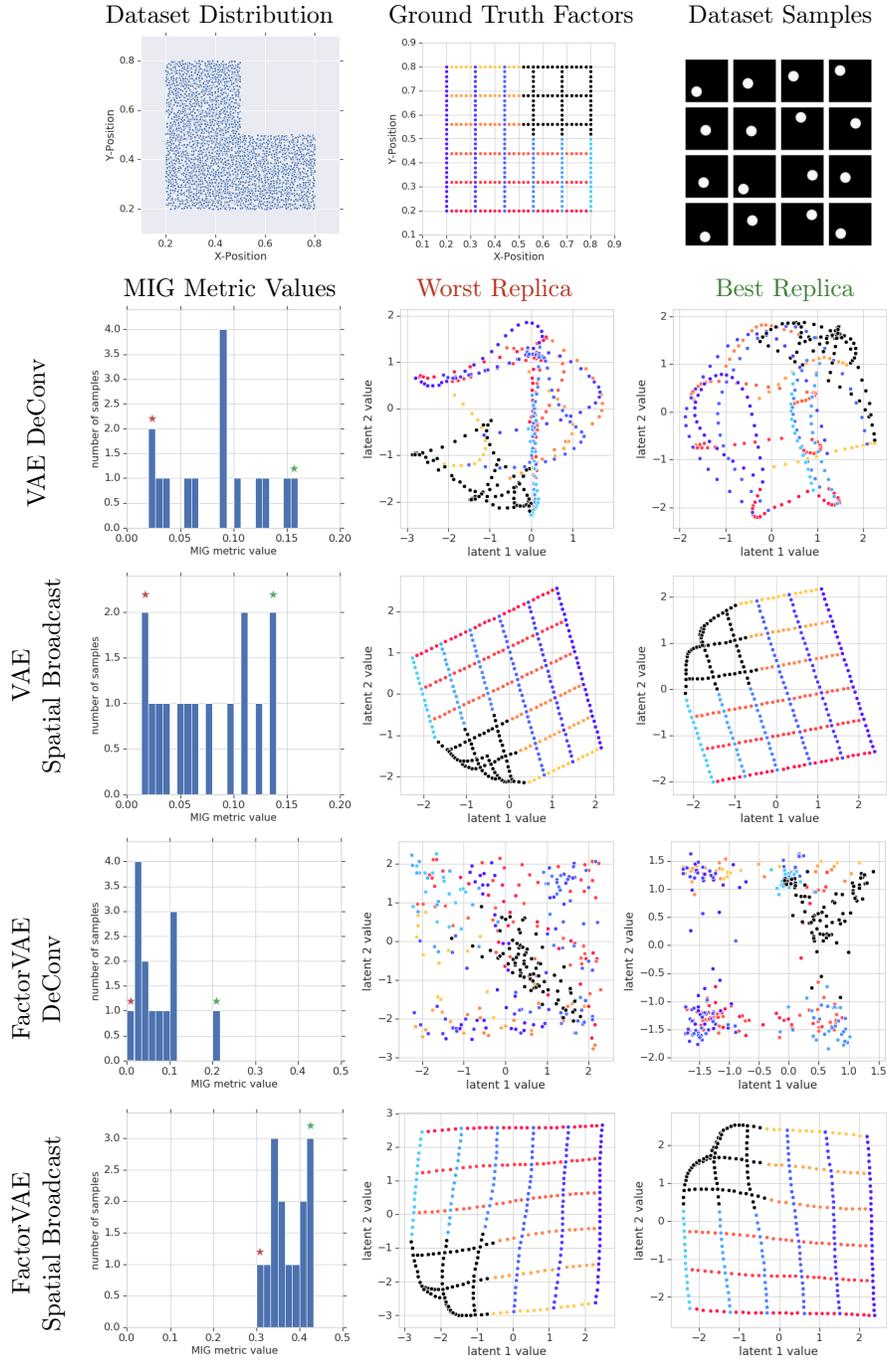

\latentgeometrytablesupp{x_y_corner}
\caption{\textbf{Latent space analysis for dependent X-Y dataset.} This is similar to Figure \ref{fig:circles:x_y_center}, except the ``hole'' in the dataset is in the corner of generative factor space rather than the middle. Hence this tests not only extrapolation in pixel space, but also extrapolation in generative factor space. As usual, the Spatial Broadcast decoder helps a lot, though in this case the extrapolation is naturally more difficult than in Figure \ref{fig:circles:x_y_center}.}
\label{fig:circles:x_y_corner}
\end{figure}

\begin{figure}[t!]
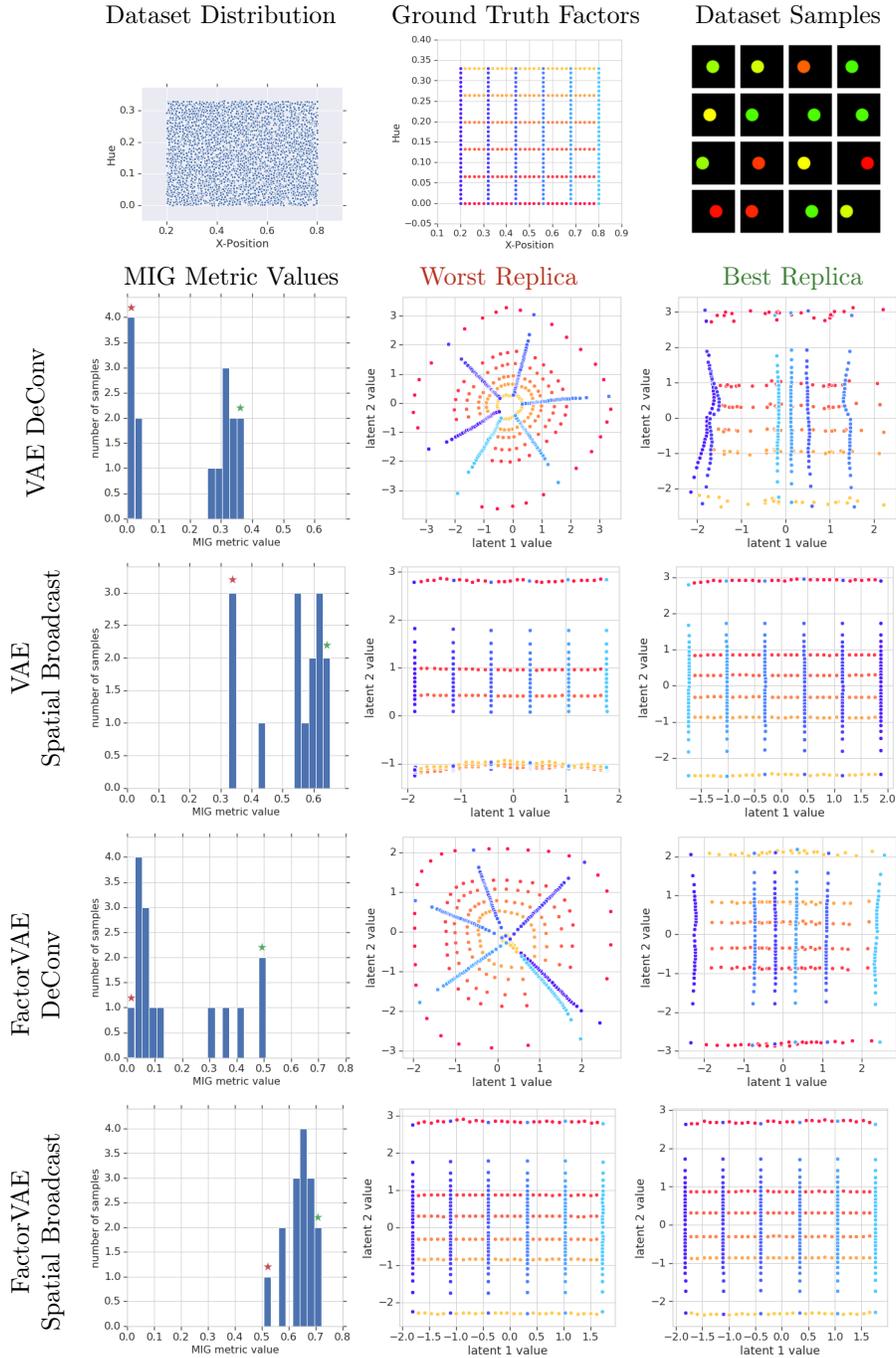

\latentgeometrytablesupp{x_h}
\caption{\textbf{Latent space analysis for independent X-H dataset.} In this dataset the circle varies in X-Position and Hue. As expected given this has less positional variation than the X-Y datasets, we see the relative improvement of the Spatial Broadcast decoder to be lower, though still quite significant. Interestingly, the representation with the Spatial Broadcast decoder is always axis-aligned and nearly linear in the positional direction, though non-linear in the hue direction. While this is not what the VAE objective is pressuring the model to do (the VAE objective would like to balance mean and variance in its latent space), we attribute this to the fact that a linear representation is much easier to compute from the coordinate channels with ReLU layers than a non-linear effect, particularly with only three convolutional ReLU layers. In a sense, the inductive bias of the architecture is overriding the inductive bias of the objective function.}
\label{fig:circles:x_h}
\end{figure}

\begin{figure}[t!]
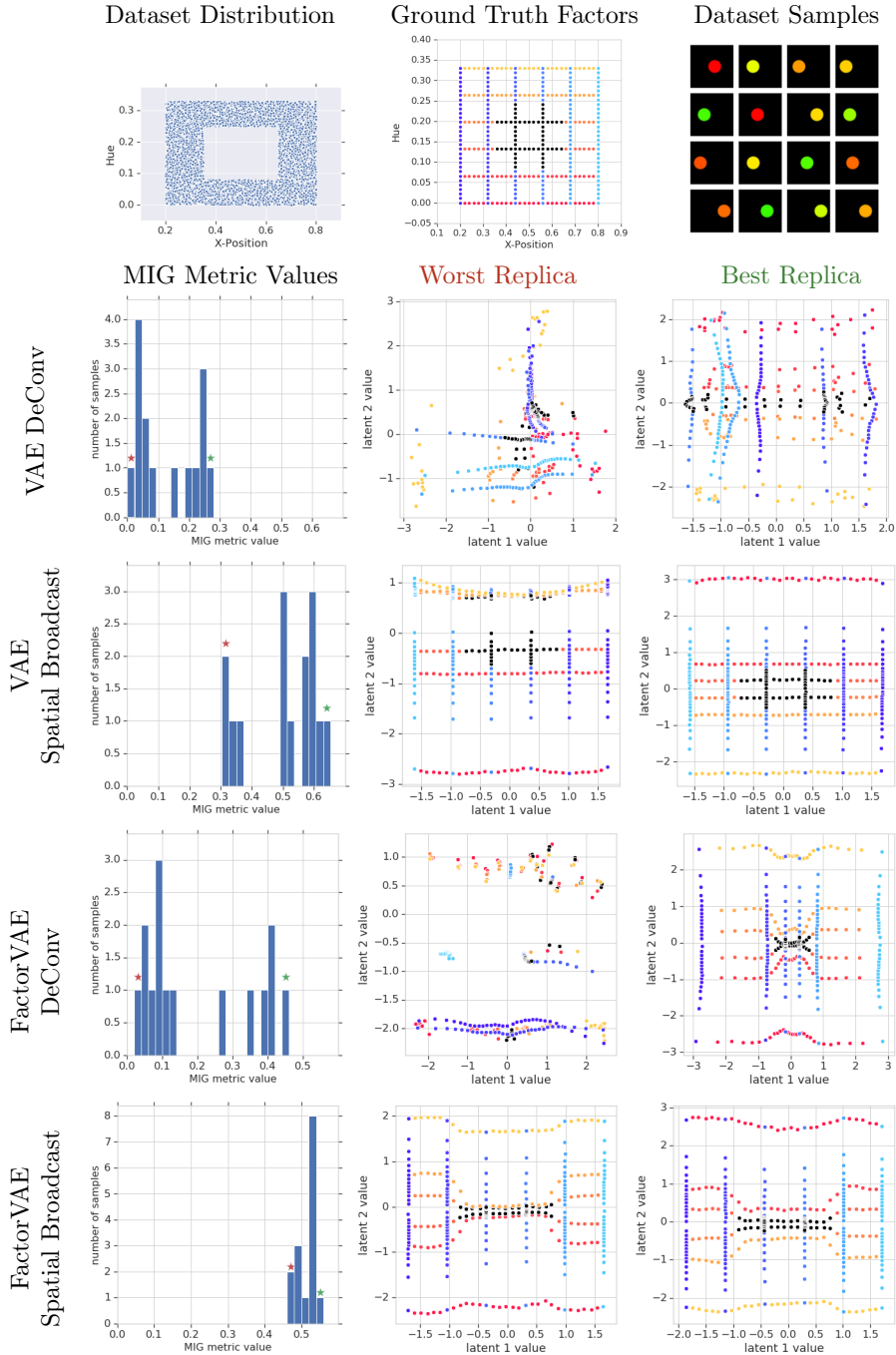

\latentgeometrytablesupp{x_h_center}
\caption{\textbf{Latent space analysis for dependent X-H dataset.} This is the same as Figure \ref{fig:circles:x_h} except the dataset has a held-out ``hole'' in the middle, hence tests the model's generalization ability. This generalization is extrapolation in pixel space yet interpolation in generative factor space. This poses a serious challenge for the DeConv decoder, and again the Spatial Broadcast decoder helps a lot. Interestingly, note the severe contraction by FactorVAE of the ``hole'' in latent space. The independence pressure in FactorVAE strongly tries to eliminate unused regions of latent space.}
\label{fig:circles:x_h_center}
\end{figure}

\begin{figure}[t!]
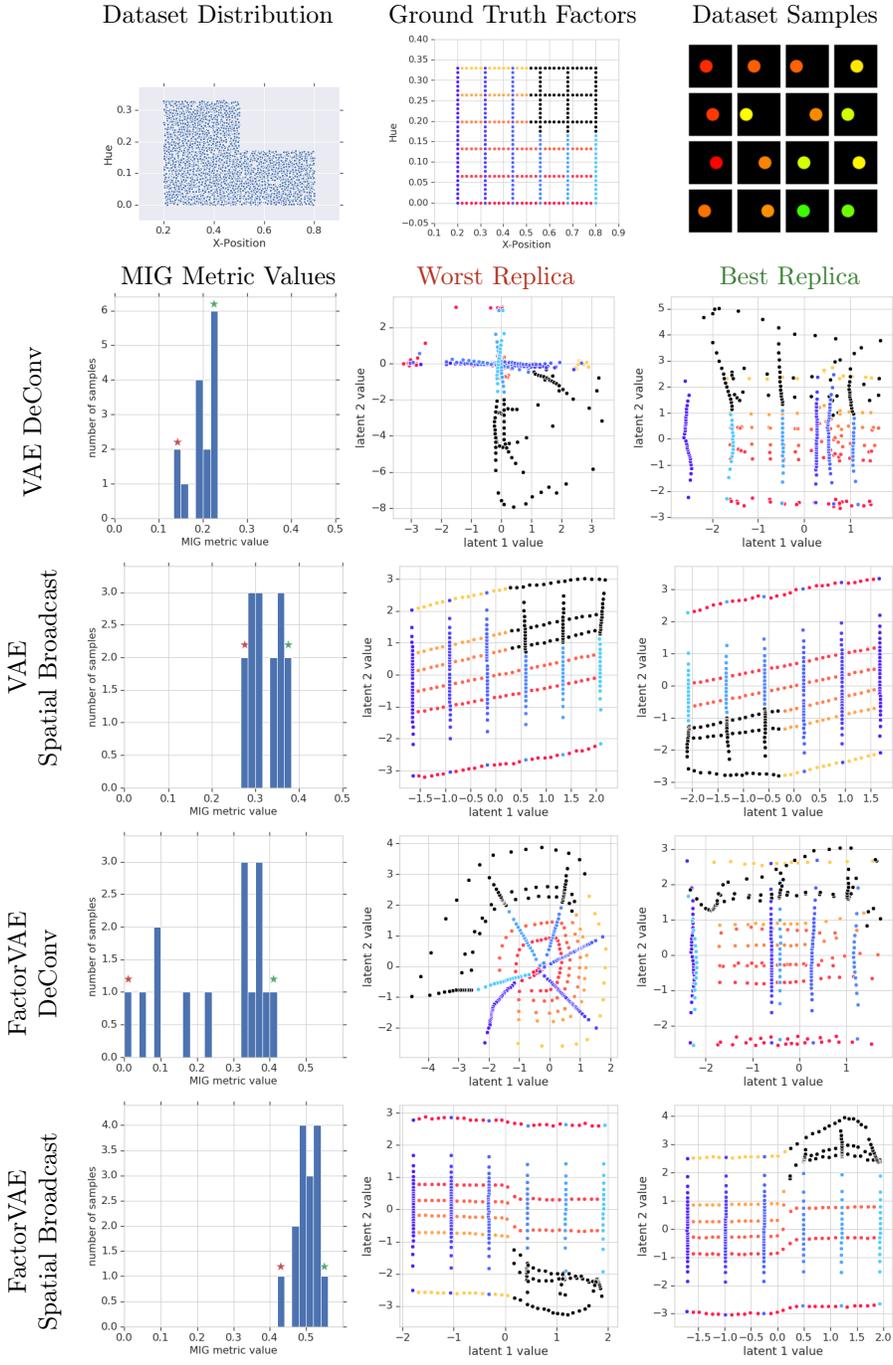

\latentgeometrytablesupp{x_h_corner}
\caption{\textbf{Latent space analysis for dependent X-H dataset.} This is the same as Figure \ref{fig:circles:x_h_center} except the held-out ``hole'' is in the corner of generative factor space, testing extrapolation in both pixel space and generative factor space. The Spatial Broadcast decoder again yields significant improvements, and as in Figure \ref{fig:circles:x_h_center} we see FactorVAE clearly sacrificing latent space geometry to remove the ``hole'' from the latent space prior distribution (see bottom row).}
\label{fig:circles:x_h_corner}
\end{figure}

\begin{figure}[t!]
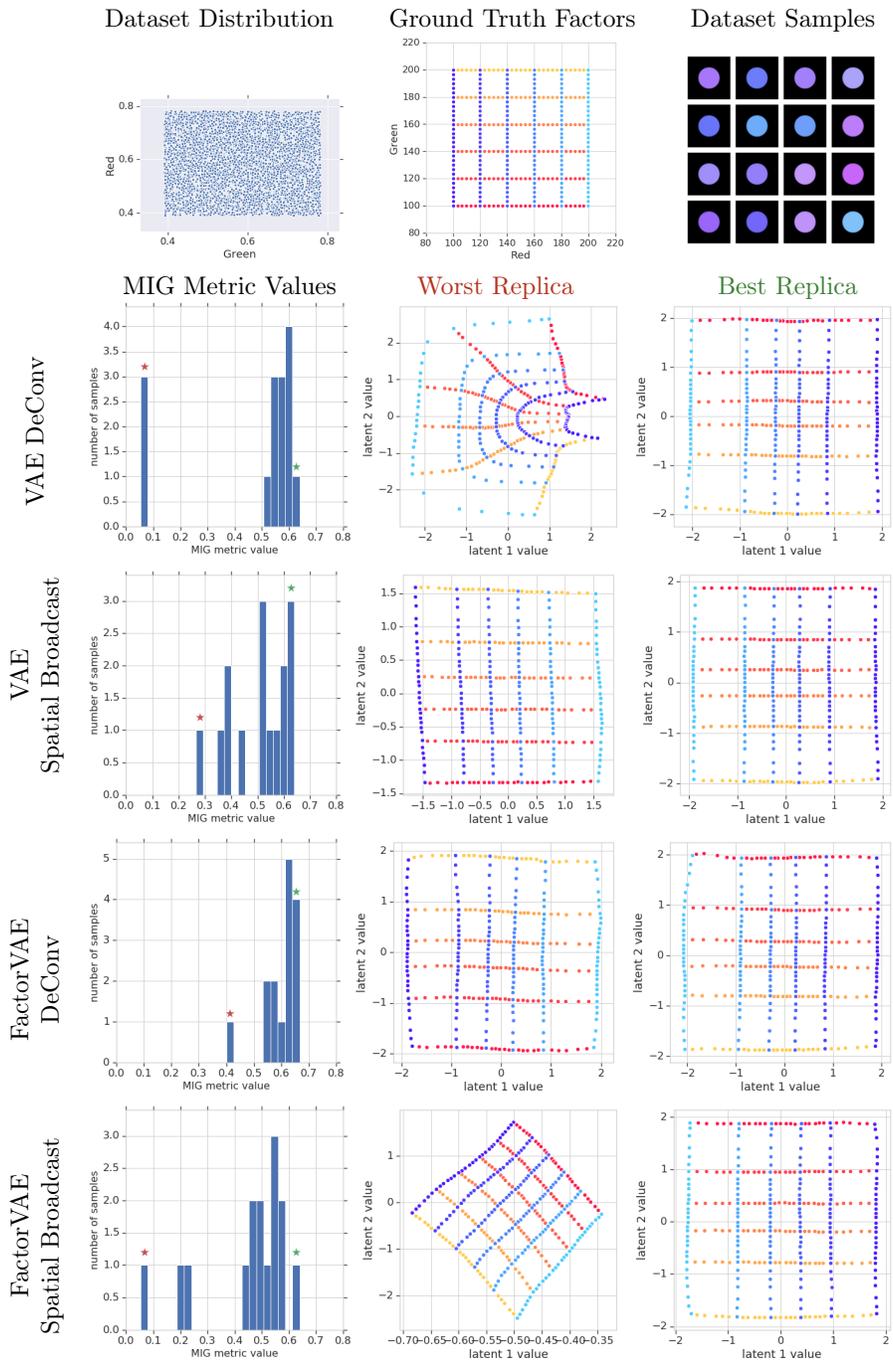

\latentgeometrytablesupp{r_g}
\caption{\textbf{Latent space analysis for independent R-G dataset.} In this dataset the circles vary only in their Redness and Greenness, not in their position. Here the benefit of the Broadcast decoder is less clear. It doesn't seem to hurt, and may help avoid a small fraction of poor seeds (3 out of 15 with MIG near zero) in a standard VAE.}
\label{fig:circles:r_g}
\end{figure}

\begin{figure}[t!]
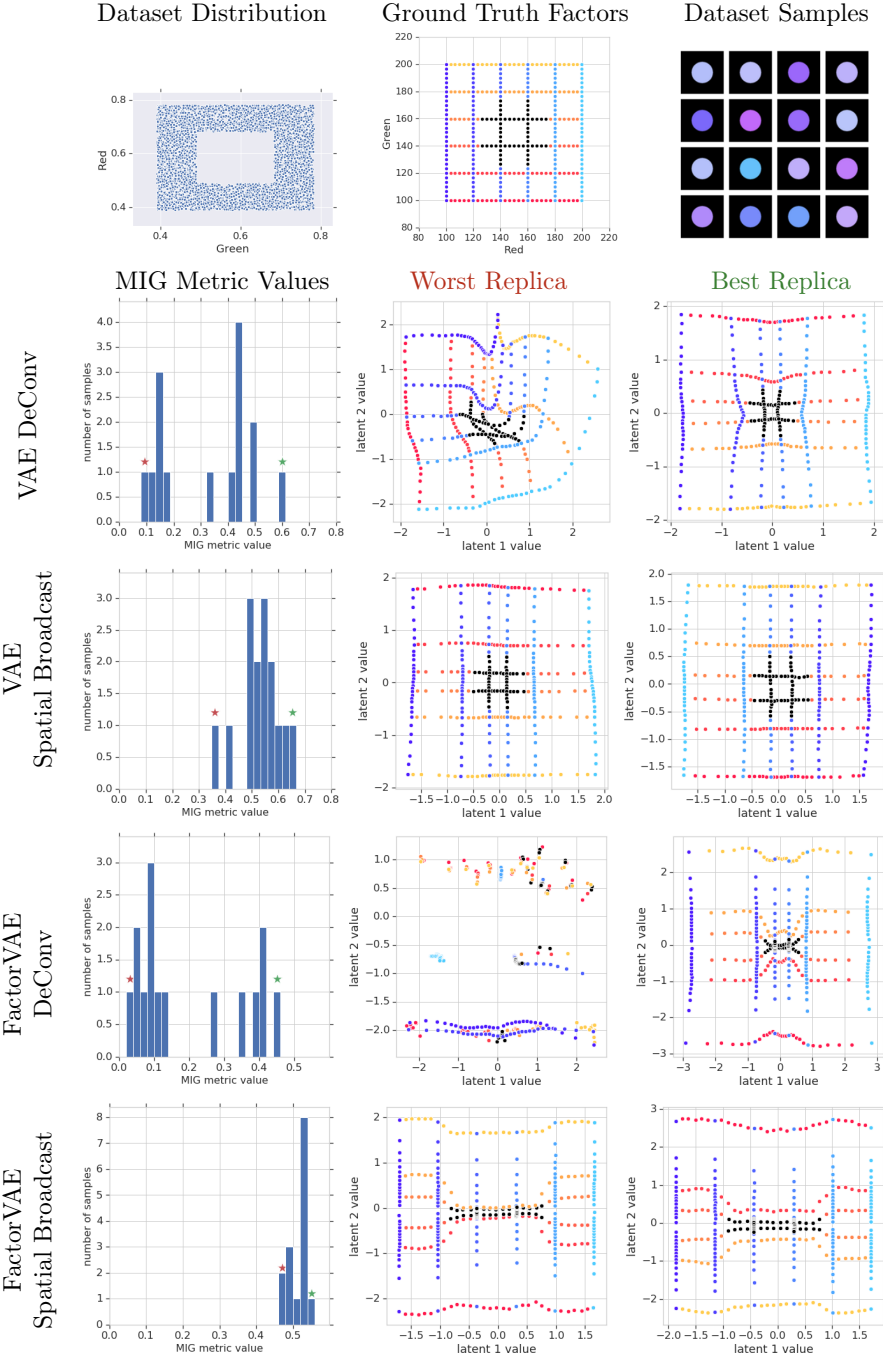

\latentgeometrytablesupp{r_g_center}
\caption{\textbf{Latent space analysis for dependent R-G dataset.} This is the same as Figure \ref{fig:circles:r_g} except with a held-out ``hole'' in the center of generative factor space, testing extrapolation in pixel space (interpolation in generative factor space). Here the benefit of the Spatial Broadcast decoder is more clear than in Figure \ref{fig:circles:r_g}. We attribute it's benefit here in part to it being a shallower network.}
\label{fig:circles:r_g_center}
\end{figure}

\begin{figure}[t!]
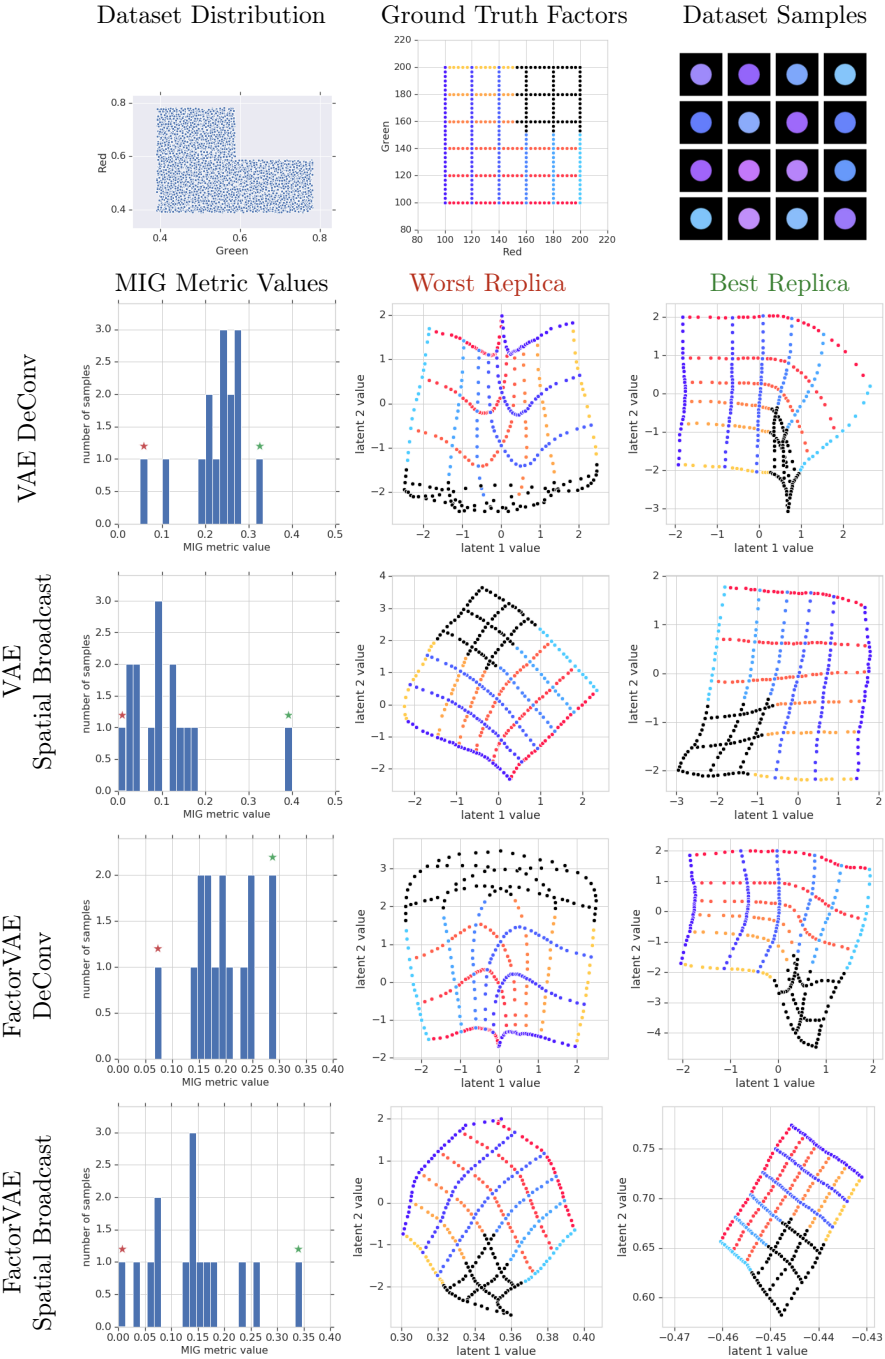

\latentgeometrytablesupp{r_g_corner}
\caption{\textbf{Latent space analysis for dependent R-G dataset.} This is the same as Figure \ref{fig:circles:r_g_center} except the ``hole'' is in the corner of generative factor space, testing extrapolation in both pixel space and generative factor space. While the Spatial Broadcast decoder seems to give rise to slightly lower MIG scores, this appears to be from rotation in the latent space. It looks like if anything the Spatial Broadcast decoder reduces warping of the latent space geometry, which is what we really care about in the representations.}
\label{fig:circles:r_g_corner}
\end{figure}

\begin{figure}[t!]
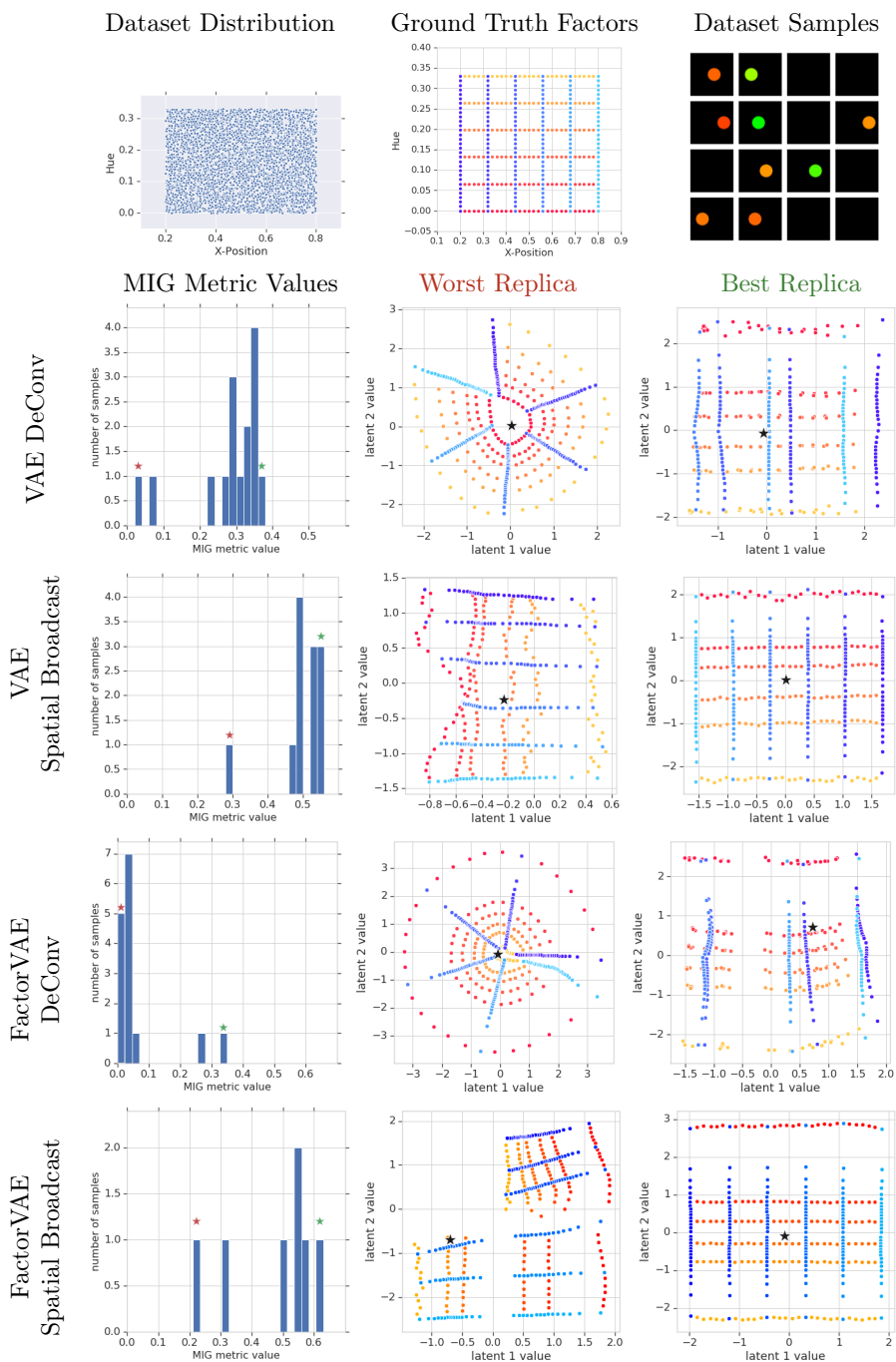

\latentgeometrytablesupp{x_h_blank}
\caption{\textbf{Latent space analysis for X-H dataset with blank images.} This is the same as Figure \ref{fig:circles:x_h} except the dataset consists half of images with no objects (entirely black images, as can be seen in the dataset samples in the top-right). This simulates the data distribution produced by the VAE of a multi-entity VAE \citep{nash17} on a dataset with a variable number of objects. Again the Spatial Broadcast decoder improves latent space geometry, according to both the MIG metric and the traversals. In the latent geometry plots, the black star represents the encoding of a blank image. We noticed that the Spatial Broadcast VAE always allocates a third relevant latent to indicate the (binary) presence or absence of an object, a very natural representation of this dataset.}
\label{fig:circles:x_h_blank}
\end{figure}

%% file: main.bbl
\begin{thebibliography}{41}
\providecommand{\natexlab}[1]{#1}
\providecommand{\url}[1]{\texttt{#1}}
\expandafter\ifx\csname urlstyle\endcsname\relax
  \providecommand{\doi}[1]{doi: #1}\else
  \providecommand{\doi}{doi: \begingroup \urlstyle{rm}\Url}\fi

\bibitem[Alemi et~al.(2017)Alemi, Poole, Fischer, Dillon, Saurous, and
  Murphy]{Alemi2017}
Alexander~A. Alemi, Ben Poole, Ian Fischer, Joshua~V. Dillon, Rif~A. Saurous,
  and Kevin Murphy.
\newblock An information-theoretic analysis of deep latent-variable models.
\newblock \emph{CoRR}, abs/1711.00464, 2017.
\newblock URL \url{http://arxiv.org/abs/1711.00464}.

\bibitem[Aubry et~al.(2014)Aubry, Maturana, Efros, Russell, and
  Sivic]{Aubry_etal_2014}
M.~Aubry, D.~Maturana, A.~Efros, B.~Russell, and J.~Sivic.
\newblock Seeing 3d chairs: exemplar part-based 2d-3d alignment using a large
  dataset of cad models.
\newblock In \emph{CVPR}, 2014.

\bibitem[Bengio et~al.(2013)Bengio, Courville, and Vincent]{Bengio_etal_2013}
Yoshua Bengio, Aaron Courville, and Pascal Vincent.
\newblock Representation learning: A review and new perspectives.
\newblock \emph{IEEE transactions on pattern analysis and machine
  intelligence}, 35\penalty0 (8):\penalty0 1798--1828, 2013.

\bibitem[Burgess et~al.(2018)Burgess, Higgins, Pal, Matthey, Watters,
  Desjardins, and Lerchner]{burgess2018}
Christopher~P. Burgess, Irina Higgins, Arka Pal, Loic Matthey, Nick Watters,
  Guillaume Desjardins, and Alexander Lerchner.
\newblock Understanding disentangling in \betavae.
\newblock \emph{arXiv}, 2018.

\bibitem[Chen et~al.(2018)Chen, Li, Grosse, and Duvenaud]{chen_2018}
Tian~Qi Chen, Xuechen Li, Roger~B. Grosse, and David~K. Duvenaud.
\newblock Isolating sources of disentanglement in variational autoencoders.
\newblock \emph{CoRR}, abs/1802.04942, 2018.
\newblock URL \url{http://arxiv.org/abs/1802.04942}.

\bibitem[Devlin et~al.(2018)Devlin, Chang, Lee, and Toutanova]{Devlin2018}
Jacob Devlin, Ming{-}Wei Chang, Kenton Lee, and Kristina Toutanova.
\newblock {BERT:} pre-training of deep bidirectional transformers for language
  understanding.
\newblock \emph{CoRR}, abs/1810.04805, 2018.
\newblock URL \url{http://arxiv.org/abs/1810.04805}.

\bibitem[Dorta et~al.(2017)Dorta, Vicente, Agapito, Campbell, Prince, and
  Simpson]{Dorta2017}
Garoe Dorta, Sara Vicente, Lourdes Agapito, Neill~D.F. Campbell, Simon Prince,
  and Ivor Simpson.
\newblock Laplacian pyramid of conditional variational autoencoders.
\newblock In \emph{Proceedings of the 14th European Conference on Visual Media
  Production (CVMP 2017)}, CVMP 2017, pages 7:1--7:9, New York, NY, USA, 2017.
  ACM.
\newblock ISBN 978-1-4503-5329-8.
\newblock \doi{10.1145/3150165.3150172}.
\newblock URL \url{http://doi.acm.org/10.1145/3150165.3150172}.

\bibitem[Eastwood and Williams(2018)]{Eastwood_Williams_2018}
Cian Eastwood and Christopher K.~I. Williams.
\newblock A framework for the quantitative evaluation of disentangled
  representations.
\newblock \emph{ICLR}, 2018.

\bibitem[Eslami et~al.(2018)Eslami, Jimenez~Rezende, Besse, Viola, Morcos,
  Garnelo, Ruderman, Rusu, Danihelka, Gregor, Reichert, Buesing, Weber,
  Vinyals, Rosenbaum, Rabinowitz, King, Hillier, Botvinick, Wierstra,
  Kavukcuoglu, and Hassabis]{Eslami2018}
S.~M.~Ali Eslami, Danilo Jimenez~Rezende, Frederic Besse, Fabio Viola, Ari~S.
  Morcos, Marta Garnelo, Avraham Ruderman, Andrei~A. Rusu, Ivo Danihelka, Karol
  Gregor, David~P. Reichert, Lars Buesing, Theophane Weber, Oriol Vinyals, Dan
  Rosenbaum, Neil Rabinowitz, Helen King, Chloe Hillier, Matt Botvinick, Daan
  Wierstra, Koray Kavukcuoglu, and Demis Hassabis.
\newblock Neural scene representation and rendering.
\newblock \emph{Science}, 360\penalty0 (6394):\penalty0 1204--1210, 2018.
\newblock ISSN 0036-8075.
\newblock \doi{10.1126/science.aar6170}.
\newblock URL \url{http://science.sciencemag.org/content/360/6394/1204}.

\bibitem[Finn et~al.(2015)Finn, Tan, Duan, Darrell, Levine, and
  Abbeel]{Finn2015}
Chelsea Finn, Xin~Yu Tan, Yan Duan, Trevor Darrell, Sergey Levine, and Pieter
  Abbeel.
\newblock Learning visual feature spaces for robotic manipulation with deep
  spatial autoencoders.
\newblock \emph{CoRR}, abs/1509.06113, 2015.
\newblock URL \url{http://arxiv.org/abs/1509.06113}.

\bibitem[Gehring et~al.(2017)Gehring, Auli, Grangier, Yarats, and
  Dauphin]{Gehring2017}
Jonas Gehring, Michael Auli, David Grangier, Denis Yarats, and Yann~N. Dauphin.
\newblock Convolutional sequence to sequence learning.
\newblock \emph{CoRR}, abs/1705.03122, 2017.
\newblock URL \url{http://arxiv.org/abs/1705.03122}.

\bibitem[Gregor et~al.(2015)Gregor, Danihelka, Graves, and
  Wierstra]{Gregor2015}
Karol Gregor, Ivo Danihelka, Alex Graves, and Daan Wierstra.
\newblock {DRAW:} {A} recurrent neural network for image generation.
\newblock \emph{CoRR}, abs/1502.04623, 2015.
\newblock URL \url{http://arxiv.org/abs/1502.04623}.

\bibitem[Gu et~al.(2016)Gu, Lu, Li, and Li]{Gu2016}
Jiatao Gu, Zhengdong Lu, Hang Li, and Victor O.~K. Li.
\newblock Incorporating copying mechanism in sequence-to-sequence learning.
\newblock \emph{CoRR}, abs/1603.06393, 2016.
\newblock URL \url{http://arxiv.org/abs/1603.06393}.

\bibitem[Higgins et~al.(2017{\natexlab{a}})Higgins, Matthey, Pal, Burgess,
  Glorot, Botvinick, Mohamed, and Lerchner]{higgins2017}
Irina Higgins, Loic Matthey, Arka Pal, Christopher Burgess, Xavier Glorot,
  Matthew Botvinick, Shakir Mohamed, and Alexander Lerchner.
\newblock \betavae: Learning basic visual concepts with a constrained
  variational framework.
\newblock \emph{ICLR}, 2017{\natexlab{a}}.

\bibitem[Higgins et~al.(2017{\natexlab{b}})Higgins, Sonnerat, Matthey, Pal,
  Burgess, Botvinick, Hassabis, and Lerchner]{higgins2017scan}
Irina Higgins, Nicolas Sonnerat, Loic Matthey, Arka Pal, Christopher~P Burgess,
  Matthew Botvinick, Demis Hassabis, and Alexander Lerchner.
\newblock Scan: learning abstract hierarchical compositional visual concepts.
\newblock \emph{arXiv preprint arXiv:1707.03389}, 2017{\natexlab{b}}.

\bibitem[Higgins et~al.(2018)Higgins, Amos, Pfau, Racani{\`{e}}re, Matthey,
  Rezende, and Lerchner]{higgins2018}
Irina Higgins, David Amos, David Pfau, S{\'{e}}bastien Racani{\`{e}}re,
  Lo{\"{\i}}c Matthey, Danilo~J. Rezende, and Alexander Lerchner.
\newblock Towards a definition of disentangled representations.
\newblock \emph{CoRR}, abs/1812.02230, 2018.
\newblock URL \url{http://arxiv.org/abs/1812.02230}.

\bibitem[Ioffe and Szegedy(2015)]{ioffe_2015}
Sergey Ioffe and Christian Szegedy.
\newblock Batch normalization: Accelerating deep network training by reducing
  internal covariate shift.
\newblock \emph{CoRR}, abs/1502.03167, 2015.
\newblock URL \url{http://arxiv.org/abs/1502.03167}.

\bibitem[Jaderberg et~al.(2015)Jaderberg, Simonyan, Zisserman, and
  Kavukcuoglu]{Jaderberg2015}
Max Jaderberg, Karen Simonyan, Andrew Zisserman, and Koray Kavukcuoglu.
\newblock Spatial transformer networks.
\newblock \emph{CoRR}, abs/1506.02025, 2015.
\newblock URL \url{http://arxiv.org/abs/1506.02025}.

\bibitem[Kim and Mnih(2017)]{Kim_Mnih_2017}
Hyunjik Kim and Andriy Mnih.
\newblock Disentangling by factorising.
\newblock \emph{arxiv}, 2017.

\bibitem[Kingma and Ba(2015)]{Kingma_Ba_2014}
Diederik~P. Kingma and Jimmy Ba.
\newblock Adam: A method for stochastic optimization.
\newblock \emph{ICLR}, 2015.

\bibitem[Kingma and Welling(2014)]{Kingma_Welling_2014}
Diederik~P. Kingma and Max Welling.
\newblock Auto-encoding variational bayes.
\newblock \emph{ICLR}, 2014.

\bibitem[Lake et~al.(2016)Lake, Ullman, Tenenbaum, and
  Gershman]{Lake_etal_2016}
Brenden~M. Lake, Tomer~D. Ullman, Joshua~B. Tenenbaum, and Samuel~J. Gershman.
\newblock Building machines that learn and think like people.
\newblock \emph{Behavioral and Brain Sciences}, pages 1--101, 2016.

\bibitem[Levine et~al.(2015)Levine, Finn, Darrell, and Abbeel]{Levine2015}
Sergey Levine, Chelsea Finn, Trevor Darrell, and Pieter Abbeel.
\newblock End-to-end training of deep visuomotor policies.
\newblock \emph{CoRR}, abs/1504.00702, 2015.
\newblock URL \url{http://arxiv.org/abs/1504.00702}.

\bibitem[Liang et~al.(2015)Liang, Wei, Shen, Yang, Lin, and Yan]{Liang2015}
Xiaodan Liang, Yunchao Wei, Xiaohui Shen, Jianchao Yang, Liang Lin, and
  Shuicheng Yan.
\newblock Proposal-free network for instance-level object segmentation.
\newblock \emph{CoRR}, abs/1509.02636, 2015.
\newblock URL \url{http://arxiv.org/abs/1509.02636}.

\bibitem[Liu et~al.(2018)Liu, Lehman, Molino, Such, Frank, Sergeev, and
  Yosinski]{liu2018}
Rosanne Liu, Joel Lehman, Piero Molino, Felipe~Petroski Such, Eric Frank, Alex
  Sergeev, and Jason Yosinski.
\newblock An intriguing failing of convolutional neural networks and the
  coordconv solution.
\newblock \emph{CoRR}, abs/1807.03247, 2018.
\newblock URL \url{http://arxiv.org/abs/1807.03247}.

\bibitem[Locatello et~al.(2018)Locatello, Bauer, Lucic, Gelly, Sch{\"o}lkopf,
  and Bachem]{Locatello_etal_2018}
Francesco Locatello, Stefan Bauer, Mario Lucic, Sylvain Gelly, Bernhard
  Sch{\"o}lkopf, and Olivier Bachem.
\newblock Challenging common assumptions in the unsupervised learning of
  disentangled representations.
\newblock \emph{arXiv preprint arXiv:1811.12359}, 2018.

\bibitem[Marcus(2018)]{marcus_2018}
Gary Marcus.
\newblock Deep learning: {A} critical appraisal.
\newblock \emph{CoRR}, abs/1801.00631, 2018.
\newblock URL \url{http://arxiv.org/abs/1801.00631}.

\bibitem[Matthey et~al.(2017)Matthey, Higgins, Hassabis, and
  Lerchner]{dsprites17}
Loic Matthey, Irina Higgins, Demis Hassabis, and Alexander Lerchner.
\newblock dsprites: Disentanglement testing sprites dataset, 2017.
\newblock URL \url{https://github.com/deepmind/dsprites-dataset/}.

\bibitem[Nash et~al.(2017)Nash, Eslami, Burgess, Higgins, Zoran, Weber, and
  Battaglia]{nash17}
Charlie Nash, Ali Eslami, Chris Burgess, Irina Higgins, Daniel Zoran, Theophane
  Weber, and Peter Battaglia.
\newblock The multi-entity variational autoencoder.
\newblock \emph{NIPS Workshops}, 2017.

\bibitem[Odena et~al.(2016)Odena, Dumoulin, and Olah]{odena2016}
Augustus Odena, Vincent Dumoulin, and Chris Olah.
\newblock Deconvolution and checkerboard artifacts.
\newblock \emph{Distill}, 2016.
\newblock \doi{10.23915/distill.00003}.
\newblock URL \url{http://distill.pub/2016/deconv-checkerboard}.

\bibitem[Parmar et~al.(2018)Parmar, Vaswani, Uszkoreit, Kaiser, Shazeer, and
  Ku]{Parmar2018}
Niki Parmar, Ashish Vaswani, Jakob Uszkoreit, Lukasz Kaiser, Noam Shazeer, and
  Alexander Ku.
\newblock Image transformer.
\newblock \emph{CoRR}, abs/1802.05751, 2018.
\newblock URL \url{http://arxiv.org/abs/1802.05751}.

\bibitem[Perez et~al.(2017)Perez, Strub, de~Vries, Dumoulin, and
  Courville]{Perez2017}
Ethan Perez, Florian Strub, Harm de~Vries, Vincent Dumoulin, and Aaron~C.
  Courville.
\newblock Film: Visual reasoning with a general conditioning layer.
\newblock \emph{CoRR}, abs/1709.07871, 2017.
\newblock URL \url{http://arxiv.org/abs/1709.07871}.

\bibitem[Reed et~al.(2016)Reed, Akata, Mohan, Tenka, Schiele, and
  Lee]{Reed2016}
Scott~E. Reed, Zeynep Akata, Santosh Mohan, Samuel Tenka, Bernt Schiele, and
  Honglak Lee.
\newblock Learning what and where to draw.
\newblock \emph{CoRR}, abs/1610.02454, 2016.
\newblock URL \url{http://arxiv.org/abs/1610.02454}.

\bibitem[Rezende et~al.(2014)Rezende, Mohamed, and Wierstra]{Rezende_etal_2014}
Danilo~Jimenez Rezende, Shakir Mohamed, and Daan Wierstra.
\newblock Stochastic backpropagation and approximate inference in deep
  generative models.
\newblock \emph{ICML}, 32\penalty0 (2):\penalty0 1278--1286, 2014.

\bibitem[Ridgeway and Mozer(2018)]{Ridgeway_Mozer_2018}
Karl Ridgeway and Michael~C Mozer.
\newblock Learning deep disentangled embeddings with the f-statistic loss.
\newblock \emph{NIPS}, 2018.

\bibitem[Ulyanov et~al.(2017)Ulyanov, Vedaldi, and Lempitsky]{Ulyanov2017}
Dmitry Ulyanov, Andrea Vedaldi, and Victor~S. Lempitsky.
\newblock Deep image prior.
\newblock \emph{CoRR}, abs/1711.10925, 2017.
\newblock URL \url{http://arxiv.org/abs/1711.10925}.

\bibitem[Vaswani et~al.(2017)Vaswani, Shazeer, Parmar, Uszkoreit, Jones, Gomez,
  Kaiser, and Polosukhin]{Vaswani2017}
Ashish Vaswani, Noam Shazeer, Niki Parmar, Jakob Uszkoreit, Llion Jones,
  Aidan~N. Gomez, Lukasz Kaiser, and Illia Polosukhin.
\newblock Attention is all you need.
\newblock \emph{CoRR}, abs/1706.03762, 2017.
\newblock URL \url{http://arxiv.org/abs/1706.03762}.

\bibitem[Watters et~al.(2017)Watters, Zoran, Weber, Battaglia, Pascanu, and
  Tacchetti]{watters_2017}
Nicholas Watters, Daniel Zoran, Theophane Weber, Peter Battaglia, Razvan
  Pascanu, and Andrea Tacchetti.
\newblock Visual interaction networks: Learning a physics simulator from video.
\newblock In \emph{Advances in Neural Information Processing Systems 30}, pages
  4539--4547. Curran Associates, Inc., 2017.
\newblock URL
  \url{http://papers.nips.cc/paper/7040-visual-interaction-networks-learning-a-physics-simulator-from-video.pdf}.

\bibitem[Watters et~al.(2019)Watters, Matthey, Borgeaud, Kabra, and
  Lerchner]{spriteworld}
Nicholas Watters, Loic Matthey, Sebastian Borgeaud, Rishabh Kabra, and
  Alexander Lerchner.
\newblock Spriteworld: A flexible, configurable reinforcement learning
  environment.
\newblock https://github.com/deepmind/spriteworld/, 2019.
\newblock URL \url{https://github.com/deepmind/spriteworld/}.

\bibitem[Wojna et~al.(2017)Wojna, Gorban, Lee, Murphy, Yu, Li, and
  Ibarz]{Wojna2017}
Zbigniew Wojna, Alexander~N. Gorban, Dar{-}Shyang Lee, Kevin Murphy, Qian Yu,
  Yeqing Li, and Julian Ibarz.
\newblock Attention-based extraction of structured information from street view
  imagery.
\newblock \emph{CoRR}, abs/1704.03549, 2017.
\newblock URL \url{http://arxiv.org/abs/1704.03549}.

\bibitem[Zhao et~al.(2015)Zhao, Mathieu, Goroshin, and LeCun]{zhao_2015}
Junbo Zhao, Michael Mathieu, Ross Goroshin, and Yann LeCun.
\newblock Stacked what-where auto-encoders.
\newblock \emph{arXiv}, abs/1506.02351, 2015.
\newblock URL \url{https://arxiv.org/abs/1506.02351}.

\end{thebibliography}
